\documentclass[preprint,12pt]{elsarticle}

\usepackage{amssymb}
\usepackage{amsmath}

\usepackage[bottom=2.5cm,top=2.5cm,left=2.5cm,right=2.5cm]{geometry} 
\usepackage{hyperref}  
\usepackage{xcolor}
\usepackage{subcaption}  

\usepackage[capitalize]{cleveref}
\crefname{section}{Sec.}{Secs.}
\Crefname{section}{Section}{Sections}
\Crefname{table}{Table}{Tables}
\crefname{table}{Tab.}{Tabs.}
\crefname{equation}{Eq.}{Eqs.}  

\usepackage[normalem]{ulem}
\useunder{\uline}{\ul}{}

\usepackage{bbding}
\usepackage{pifont}
\usepackage{wasysym}
\usepackage{amssymb}

\usepackage{multirow}                 

\journal{ISPRS Journal of Photogrammetry and Remote Sensing}

\begin{document}

\begin{frontmatter}



\title{Enhancing 3D LiDAR Segmentation by Shaping Dense and Accurate 2D Semantic Predictions} 

\author[label1,label2]{Xiaoyu Dong}
\author[label1]{Tiankui Xian}
\author[label1,label2]{Wanshui Gan}  
\author[label1,label2]{Naoto Yokoya} 
\affiliation[label1]{organization={The University of Tokyo},
            city={Tokyo},
            country={Japan}}

\affiliation[label2]{organization={RIKEN AIP},
            city={Tokyo},
            country={Japan}}



\begin{abstract}
Semantic segmentation of 3D LiDAR point clouds is important in urban remote sensing for understanding real-world street environments.
This task, by projecting LiDAR point clouds and 3D semantic labels as sparse maps, can be reformulated as a 2D problem. 
However, the intrinsic sparsity of the projected LiDAR and label maps can result in sparse and inaccurate intermediate 2D semantic predictions, which in return limits the final 3D accuracy.
To address this issue, we enhance this task by shaping dense and accurate 2D predictions. 
Specifically, we develop a multi-modal segmentation model, MM2D3D. 
By leveraging camera images as auxiliary data, we introduce cross-modal guided filtering to overcome label map sparsity by constraining intermediate 2D semantic predictions with dense semantic relations derived from the camera images; and we introduce dynamic cross pseudo supervision to overcome LiDAR map sparsity by encouraging the 2D predictions to emulate the dense distribution of the semantic predictions from the camera images. 
Experiments show that our techniques enable our model to achieve intermediate 2D semantic predictions with dense distribution and higher accuracy, which effectively enhances the final 3D accuracy.  
Comparisons with previous methods demonstrate our superior performance in both 2D and 3D spaces.
\end{abstract}



\begin{keyword}
LiDAR Point Clouds \sep 3D Semantic Segmentation \sep 2D Semantic Segmentation \sep Camera-LiDAR Fusion



\end{keyword}

\end{frontmatter}




\section{Introduction}
\label{sec:intro}

\begin{figure}[!ht]
  \centering
  \begin{subfigure}{0.5\linewidth} 
    \centering
    \begin{minipage}{0.3\linewidth} 
        \centering
        \includegraphics[width=1\textwidth]{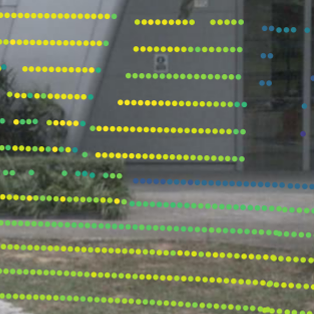}
        {\scriptsize Inputs}
    \end{minipage}
    \begin{minipage}{0.3\linewidth}  
        \centering
        \includegraphics[width=1\textwidth]{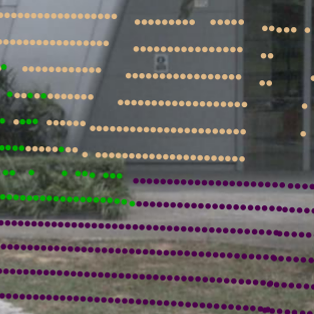}
        {\scriptsize P3D Label}
    \end{minipage}
    \begin{minipage}{0.3\linewidth} 
        \centering
        \includegraphics[width=1\textwidth]{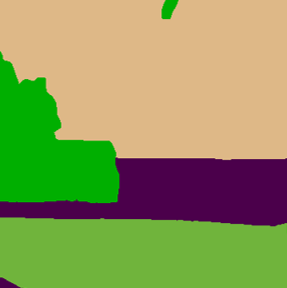}
        {\scriptsize 2D Label}
    \end{minipage}
    \begin{minipage}{0.3\linewidth}  
        \centering
        \includegraphics[width=1\textwidth]{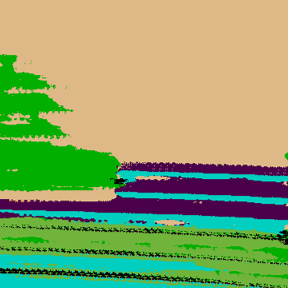}
        {\scriptsize PMF~\cite{iccv21_pmf}}
    \end{minipage}
    \begin{minipage}{0.3\linewidth}  
        \centering
        \includegraphics[width=1\textwidth]{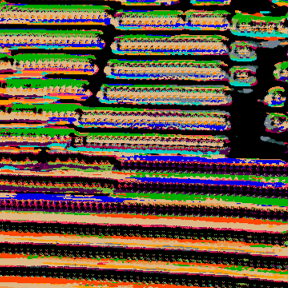}
        {\scriptsize Baseline}
    \end{minipage}
    \begin{minipage}{0.3\linewidth}  
        \centering
        \includegraphics[width=1\textwidth]{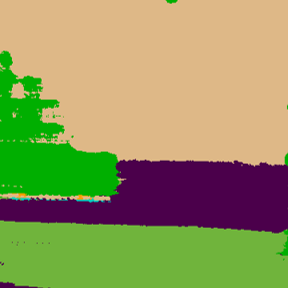}
        {\scriptsize MM2D3D (Ours)}
    \end{minipage}
    \caption{ } 
    \label{fig1a}
  \end{subfigure}
  \begin{subfigure}{0.45\linewidth}   
    \includegraphics[width=1\linewidth]{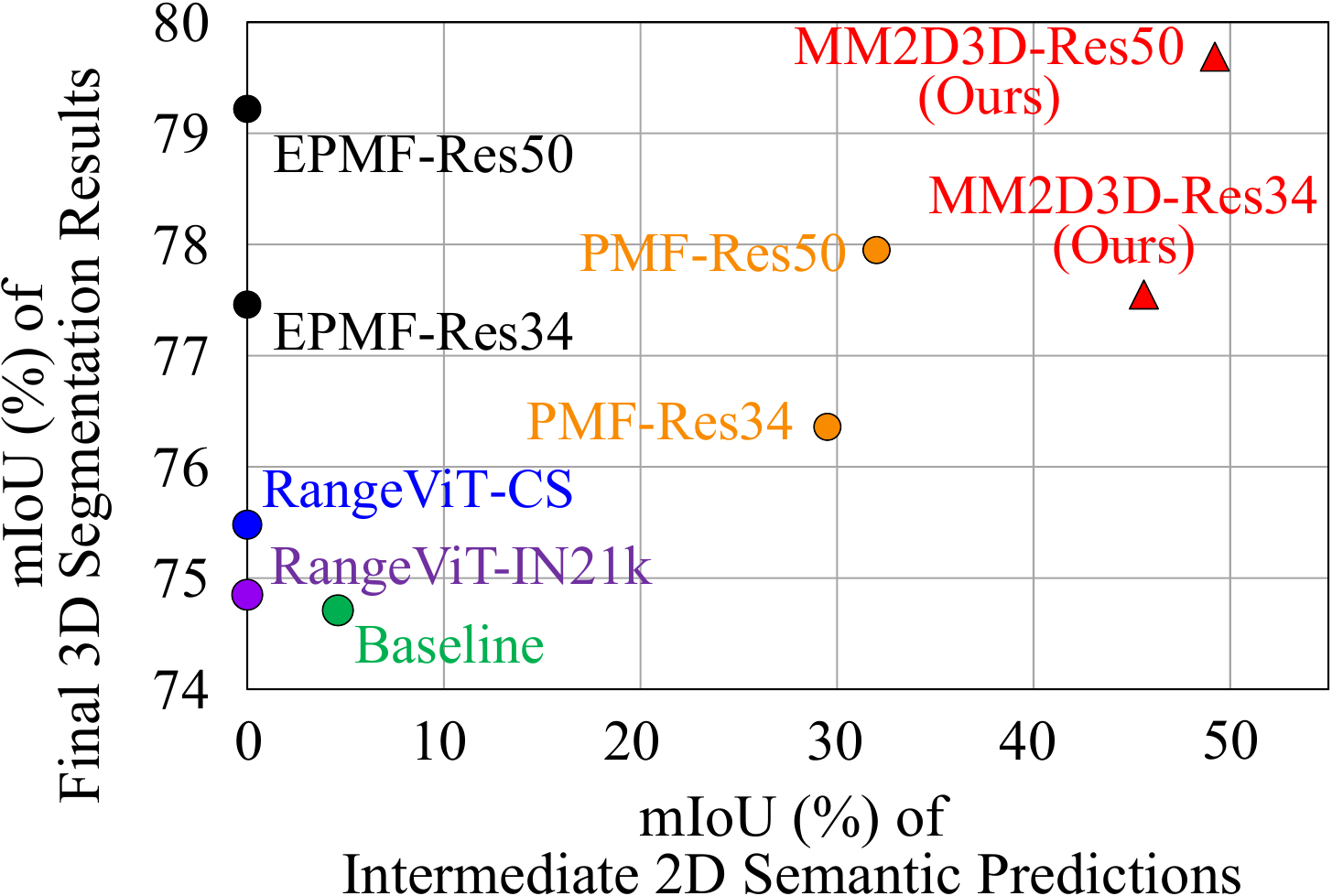}
    \vspace{0.01cm}
    \caption{} 
    \label{fig1b}
  \end{subfigure}
  \caption{ 
  (a) Intermediate 2D semantic predictions.
  Inputs are LiDAR map and camera image. 
  P3D Label indicates label map (shown stacked on camera image) and is used as supervision for training. 
  2D Label is only used for evaluation.
  (b)~Comparison of 2D and 3D accuracy on our nuScenes2D3D test set. 
  The 2D accuracy of RangeViT-CS~\cite{cvpr23_rangevit}, RangeViT-IN21k~\cite{cvpr23_rangevit}, and EPMF~\cite{pami24_epmf} is not available because their intermediate 2D predictions and 2D labels are from different projection views or misaligned.
  }
  \label{fig1}
\end{figure}


Semantic segmentation of 3D LiDAR point clouds, which aims to predict class information for each point in the point clouds to understand the different parts and regions of surrounding environment, plays a crucial role in urban remote sensing. 
Due to the sparse and irregular distribution of point clouds, which causes difficulties in capturing object details, camera images are often used as auxiliary data to improve the accuracy and robustness of LiDAR segmentation~\cite{iccv23_see_more,pami22_xmuda}. 

To address point-image heterogeneity, projection-based methods~\cite{cvpr23_rangevit,iccv21_pmf,wacv20_fuseseg,itsc19_rgbal,pami24_epmf} reformulate this task as a 2D problem, where LiDAR point clouds and 3D semantic labels are projected as sparse maps to better integrate the structures and semantics from camera images and produce intermediate semantic predictions in 2D space. 
Final 3D segmentation results can be obtained by remapping corresponding points in the intermediate 2D semantic predictions back to the point clouds.   
However, due to intrinsic sparsity issues, the intermediate 2D predictions can suffer from significant sparsity and inaccuracy (see~\cref{fig1a}):
(\uppercase\expandafter{\romannumeral1})~The input LiDAR maps have a sparse spatial distribution, leading to sparsity (``black holes'') in the 2D predictions.
(\uppercase\expandafter{\romannumeral2})~The supervision label maps are also sparse, resulting in inaccuracies in unlabeled regions. 
Previous projection-based methods focus on developing advanced network structures and training strategies to better process learned LiDAR features or integrate information from camera images, lacking considerations on these issues.  
As a result, they tend to produce sparse and inaccurate intermediate 2D predictions, which can limit the final 3D accuracy because, in the projection-based setting: 3D results are derived from 2D predictions, and the inference on projected points heavily relies on the semantics of their neighboring pixels in 2D space.
In other words, achieving dense and accurate intermediate 2D predictions is important to the final 3D accuracy. 
However, this remains a challenge due to the intrinsic sparsity issues.

This paper makes an attempt to enhance 3D LiDAR segmentation in the projection-based setting by shaping dense and accurate 2D semantic predictions.
Specifically, we develop a multi-modal segmentation model, MM2D3D, as illustrated in~\cref{fig2_model}.
By leveraging camera images as auxiliary data, we introduce two techniques—cross-modal guided filtering and dynamic cross pseudo supervision—to address the intrinsic sparsity issues.
The cross-modal guided filtering (\cref{fig3_filtering}) aims to overcome the label map sparsity and increase prediction accuracy in unlabeled regions by constraining the intermediate 2D semantic predictions with dense semantic relations derived from camera images. 
The dynamic cross pseudo supervision (\cref{fig2_model}) aims to overcome the LiDAR map sparsity and densify the intermediate 2D predictions by encouraging them to emulate the dense distribution of the semantic predictions from camera images\footnote{In this paper, \textit{intermediate 2D semantic predictions} specifically refer to the semantic predictions obtained from LiDAR maps, although semantic predictions from camera images are also produced in 2D space.}. 
It is demonstrated that: 
(\uppercase\expandafter{\romannumeral1})~Our techniques enable dense and accurate 2D semantic predictions for sparse LiDAR maps, even with only sparse label maps as supervision for training~(\cref{fig1a}). 
(\uppercase\expandafter{\romannumeral2})~Our improvements in 2D space effectively enhance the final 3D accuracy (\cref{fig1b}).

\textbf{Contributions:} 
(1) We enhance 3D LiDAR segmentation in the projection-based setting by shaping dense and accurate 2D semantic predictions. 
(2) We demonstrate that our two techniques, by leveraging cross-modal semantic guidance and encouraging dynamic cross-mimicking, effectively address the sparsity issues and enhance the final 3D accuracy.
(3) Based on nuScenes~\cite{cvpr20_nuscenes}, we introduce nuScenes2D3D, 
which provides both 2D and 3D semantic labels for camera-LiDAR data, to support our research and contribute to the community.
(4) We comprehensively analyze our model and compare it with state-of-the-art methods, showcasing its effectiveness and superiority in both 2D and 3D spaces.


\section{Related Work}
\label{sec:related}

\subsection{3D LiDAR Semantic Segmentation}

3D LiDAR semantic segmentation methods can be categorized into point-based, voxel-based, and projection-based methods, based on how they process point clouds.
Point-based methods~\cite{cvpr17_pointnet,nips17_pointnet++,iccv19_kpconv,cvpr20_randlanet,cvpr20_pointasnl,cvpr21_realpc,wu2022ptv2,wu2024ptv3} perform semantic segmentation by directly processing 3D point clouds. 
These methods may face heavy computation, particularly when sampling and aggregating disordered neighbor points in large-scale scenes.
Voxel-based methods~\cite{eccv20_spvnas,cvpr21_cylindr,iccv21_rpvnet,cvpr21_af2s3net,eccv22_self_distill,cvpr23_mseg3d,Ye_2023_AAAI,Chen_2023_ICCV} convert point clouds into 3D voxels and then apply 3D convolutions to these voxels. 
This conversion complicates the integration of information from camera images, due to the large domain gap between 3D voxels and 2D images.
Projection-based methods transform point clouds into 2D maps in range view~\cite{itsc19_rgbal,iros19_rangenet++,isvc20_salsanext,eccv20_squeezesegv3,wacv20_fuseseg,cvpr23_rangevit,iccv23_rangeformer} or perspective (camera) view~\cite{iccv21_pmf,pami24_epmf}, 
allowing for flexible and effective utilization of camera images to enhance the segmentation accuracy. 
Our work adopts the perspective-view projection setting to more effectively leverage information from camera images. 


\subsection{Driving 3D LiDAR Segmentation in 2D Space}  

3D LiDAR segmentation can be driven in 2D space through the projection and remapping of the point clouds.
A family of methods~\cite{3dv21_2d3dnet,cvpr22_slidr,nips23_seal} distills 2D image segmentation models to learn 3D LiDAR segmentation without using ground-truth 3D labels.  
In~\cite{3dv21_2d3dnet}, LiDAR segmentation models are trained using pseudo labels derived from 2D segmentation results of labeled images. 
\cite{cvpr22_slidr} proposes a pretraining method for LiDAR segmentation by distilling self-supervised pretrained image representations into 3D models. 
\cite{nips23_seal} introduces a self-supervised LiDAR segmentation model by leveraging vision foundation models and the 2D-3D correspondence between LiDAR and camera sensors. 
However, the performance of such methods is often constrained by their unsupervised learning manner.
Another family of methods, namely projection-based LiDAR segmentation methods~\cite{isvc20_salsanext,iccv23_rangeformer,cvpr23_rangevit,iccv21_pmf}, trains 2D segmentation models by using projected LiDAR and label maps.
Techniques such as dual-stream networks with perception-aware multi-sensor fusion~\cite{iccv21_pmf} and advanced pretraining strategies based on image classification and segmentation tasks~\cite{cvpr23_rangevit} have been developed to facilitate this process.
Despite these advancements, their intermediate 2D semantic predictions suffer from sparsity and inaccuracy due to the intrinsic sparsity of LiDAR and label maps, which can still limit the final 3D accuracy. 


\subsection{Learnable Image Filtering}

Learnable image filtering operations, such as local~\cite{icip94_local,pami13_guided_filtering}, non-local~\cite{nips18_nlrn,cvpr21_nonlocal}, and tree~\cite{pami15_stereotree,cvpr16_salienttree,nips19_ltf} filters, are developed in vision tasks to suppress and/or extract content of interests in images.  
In image segmentation, non-local~\cite{cvpr18_nlneural,iccv19_asymmetric} and tree filters~\cite{nips19_ltf,cvpr22_tel} are employed to model class dependencies and semantic relations among pixels.   
\cite{nips19_ltf} demonstrates that minimum spanning tree~\cite{pami15_stereotree} outperforms non-local operations in capturing object structures and intra-class dependencies.  

We aim to enhance 3D LiDAR segmentation in the projection-based setting by shaping dense and accurate 2D semantic predictions.
Our task is much more challenging than image segmentation due to: 
Both input LiDAR maps and supervision label maps are sparse, and there exist spatial misalignment and information discrepancy between LiDAR maps and camera images.
To address these issues, based on a minimum spanning tree, we propose a novel cross-modal guided filtering technique that leverages dense semantic relations derived from \textit{low-level features of camera images} to constrain intermediate 2D semantic predictions.
In \cref{subsec:ablation}, we show that our guided filtering effectively overcomes label map sparsity, increases accuracy in unlabeled regions, and addresses the misalignment and discrepancy issues.


\subsection{Cross Supervision}  

Cross supervision strategies, which encourage a source to mimic the properties of a target within or across domains, are emerging in various research topics~\cite{cvpr22_cmpl,cvpr21_cps,cvpr22_u2pl,eccv22_mmsr}.
In cross-modal super-resolution, Dong \textit{et al.}~\cite{eccv22_mmsr} propose a mutual modulation strategy where two different image modalities are encouraged to emulate the spatial resolution or modality characteristics of each other. 
In semi-supervised semantic segmentation, Chen \textit{et al.}~\cite{cvpr21_cps} introduce cross pseudo supervision to enforce consistency between student and teacher networks.
\textcolor{black}{In multi-modal semantic segmentation, cross supervision can be developed to enhance the prediction of one modality by benefiting from the complementary information in the other~\cite{cvpr20_xmuda,iccv21_pmf,aaai20_medical,cvpr21_abmdernet,wacv24_smmcl}.}  

In our task, we introduce a dynamic cross pseudo supervision strategy to address LiDAR map sparsity by encouraging intermediate 2D semantic predictions to emulate the dense distribution of camera semantic predictions. 
Our strategy differs from others by:   
(\uppercase\expandafter{\romannumeral1}) It considers the reliability of predictions in a dynamic manner. 
(\uppercase\expandafter{\romannumeral2})~It imposes an additional constraint on unlabeled regions, in conjunction with our guided filtering.
Experiments in~\cref{subsec:ablation} demonstrate the effectiveness of our strategy and the necessity of jointly applying our techniques.


\begin{figure}[!t]
  \centering
  \includegraphics[width=0.9\linewidth]{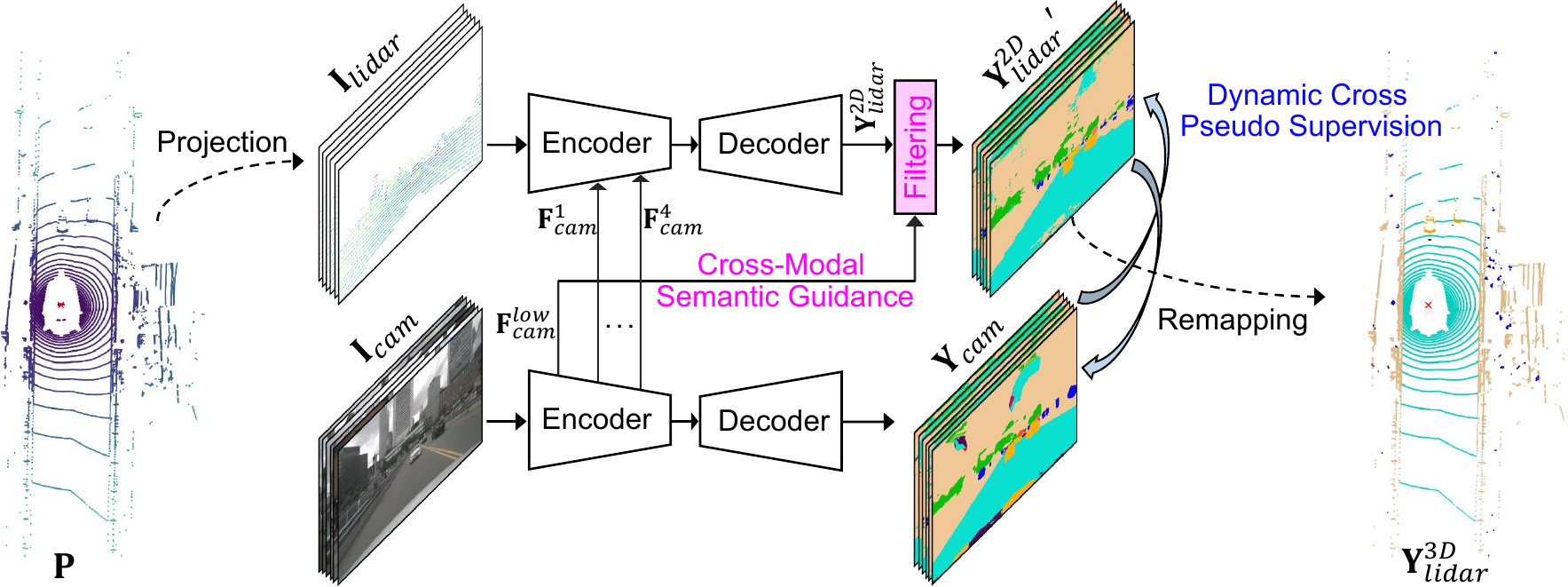}
  \caption{
  An illustration of our MM2D3D model.
  The cross-modal guided filtering constrains intermediate 2D semantic predictions with dense semantic relations derived from camera images to increase accuracy in unlabeled regions. 
  The dynamic cross pseudo supervision encourages the intermediate 2D semantic predictions to emulate the dense distribution of camera semantic predictions. 
  }
  \label{fig2_model}
\end{figure}

\section{Method}
\label{sec:method}

This section first provides an overview of our model and then details our introduced two techniques.

\subsection{Model Overview}
\label{subsec:model_overview}

Our MM2D3D model is illustrated in~\cref{fig2_model}.
Given a LiDAR point cloud $\textbf{P}\in\mathbb{R}^{N\times 4}$ and corresponding camera images $\textbf{I}^{i}_{cam}\in\mathbb{R}^{3\times H\times W}$,  
we first project $\textbf{P}$ into LiDAR maps $\textbf{I}^{i}_{lidar}\in\mathbb{R}^{5\times H\times W}$ in the perspective (camera) views.
Here, $N$ is the number of points, with each point having 3D coordinates $(x,y,z)$ and reflectance $r$.
$i$ is view index\footnote{In~\cref{fig2_model} and throughout this paper, we omit the view index for simplicity.}.  
In $\textbf{I}_{lidar}$, each pixel corresponding to a projected point contains channel values $(d,x,y,z,r)$, where $d=\sqrt{x^2 + y^2 + z^2}$ is the range value of the point. All other pixels are initialized to zero.  

We use two encoders to process $\textbf{I}_{lidar}$ and $\textbf{I}_{cam}$, extracting multi-modal features $\textbf{F}_{lidar}^s\in\mathbb{R}^{C_{lidar}^s\times H^s\times W^s}$ and $\textbf{F}_{cam}^s\in\mathbb{R}^{C_{cam}^s\times H^s\times W^s}$, where $s=1,2,3,4$ denotes the encoder stage.  
To facilitate LiDAR map segmentation with camera features, we update {$\textbf{F}_{lidar}^s$ as:   
\begin{equation}
  {\textbf{F}^{s}_{lidar}}' = \textbf{F}_{lidar}^s + f_{map}(\textbf{F}_{cam}^s),
  \label{eq_fusion}
\end{equation}
where $f_{map}(\cdot)$ is a mapping layer that transforms $C_{cam}^s$ to $C_{lidar}^s$.

The two decoders produce $\textbf{Y}_{lidar}^{2D}\in\mathbb{R}^{N_{cls}\times H\times W}$ and $\textbf{Y}_{cam}\in\mathbb{R}^{N_{cls}\times H\times W}$,  
which are the semantic predictions for the LiDAR map and the camera image, respectively. 
Here, $N_{cls}$ is the number of semantic classes.
After the LiDAR decoder, our cross-modal guided filtering updates $\textbf{Y}_{lidar}^{2D}$ to generate the intermediate 2D semantic prediction ${\textbf{Y}_{lidar}^{2D}}'\in\mathbb{R}^{N_{cls}\times H\times W}$.

During training, 
${\textbf{Y}_{lidar}^{2D}}'$ and $\textbf{Y}_{cam}$ are supervised with a sparse label map using multi-class focal loss~\cite{iccv17_focal} and Lovász-softmax loss~\cite{cvpr18_lovasz}: $\mathcal{L}_{foc}^{lidar}$, $\mathcal{L}_{lov}^{lidar}$, $\mathcal{L}_{foc}^{cam}$, and $\mathcal{L}_{lov}^{cam}$.
Additionally, they are regularized through our dynamic cross pseudo supervision. 
The final 3D segmentation result, 
$\textbf{Y}_{lidar}^{3D}\in\mathbb{R}^{N}$, is obtained by remapping the corresponding points from ${\textbf{Y}_{lidar}^{2D}}'$ to the point cloud.


\begin{figure*}[t]
  \centering
  \begin{subfigure}{0.35\linewidth}
    \includegraphics[width=1\linewidth]{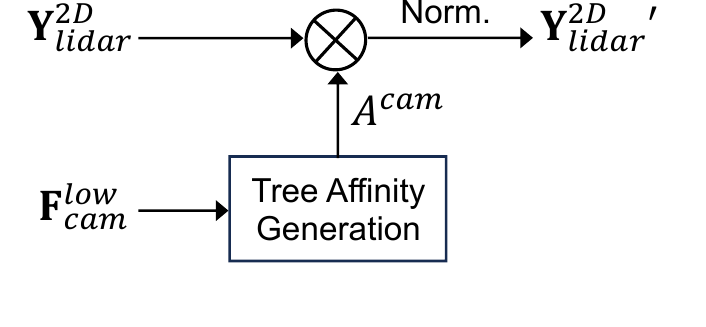}
    \caption{}
    \label{fig3a_filtering}
  \end{subfigure}
  \begin{subfigure}{0.6\linewidth}
    \includegraphics[width=1\linewidth]{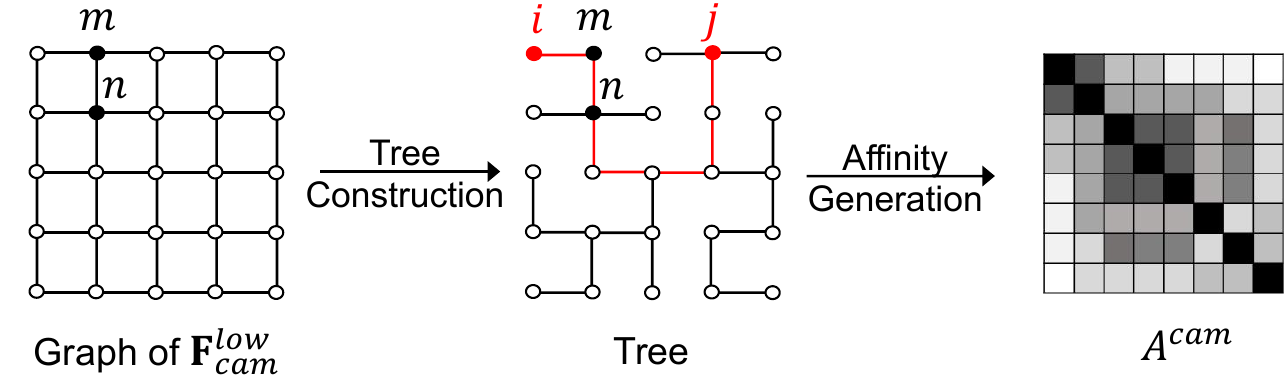}
    \caption{}
    \label{fig3b_affinity}
  \end{subfigure}
  \caption{
  (a) Our cross-modal guided filtering.
  (b) Tree affinity generation: In the 4-connected planar graph, each vertex has four neighbors. The minimum spanning tree includes all vertices in the graph, ensuring connectivity. We generate the affinity matrix based on the distances between vertices in the tree.
  }
  \label{fig3_filtering}
\end{figure*}

 
\subsection{Cross-Modal Guided Filtering}
\label{subsec:filtering}

We aim to enhance 3D LiDAR segmentation in the projection-based setting by achieving dense and accurate intermediate 2D semantic predictions.
One challenge is the sparsity of supervision label maps, which causes a lack of semantic constraints in unlabeled regions.
To address this, we introduce novel cross-modal guided filtering~(\cref{fig3a_filtering}).

The key idea of our guided filtering is to leverage dense semantic relations from the low-level structures and contexts in the camera image to constrain unlabeled regions in $\textbf{Y}{lidar}^{2D}$. 
Specifically, we first extract a low-level feature $\textbf{F}_{cam}^{low}\in\mathbb{R}^{32\times H\times W}$ after the first convolution layer in the camera encoder. 
Then, we represent $\textbf{F}_{cam}^{low}$ using a minimum spanning tree and generate an affinity matrix to model pixel dependencies and derive dense semantic relations.

As shown in~\cref{fig3b_affinity}, we construct a 4-connected planar graph $G=(V,E)$ on $\textbf{F}_{cam}^{low}$, where the vertex set $V$ includes all pixels in $\textbf{F}_{cam}^{low}$ and the edges between adjacent vertices form the edge set $E$.
The edge weight (dissimilarity) between adjacent vertices $m$ and $n$ is computed as: 
\begin{equation} 
  w_{m,n} = w_{n,m} = {\big\|} \textbf{F}_{cam}^{low}(m) - \textbf{F}_{cam}^{low}(n) {\big\|}_2.
  \label{eq_edgeweight}
\end{equation}
Using the algorithm from~\cite{1983_mst}, we construct a minimum spanning tree by removing the edge with the largest dissimilarity from $E$, while maintaining the connectivity of $G$. 
This models the dependencies among pixels in $\textbf{F}_{cam}^{low}$. 
The distance map of the tree, which measures the distance between any two vertices $i$ and $j$, is defined as a sum of dissimilarities along the path connecting them: 
\begin{equation}
  D_{i,j} = D_{j,i} = \sum_{(m,n)\in\textbf{E}_{i,j}} w_{m,n}.
  \label{eq_distance}
\end{equation}
Here, $\textbf{E}_{i,j}$ is the set of edges connecting vertex $i$ to $j$. 
Converting this distance map into a positive scalar value, we obtain:
\begin{equation}
  A^{cam} = {\rm exp}(-D).
  \label{eq_affinity}
\end{equation}
$A^{cam}$ is the affinity matrix that captures the dense semantic relations (similarities and dissimilarities) among vertices in the tree of $\textbf{F}_{cam}^{low}$.  

The last step of our guided filtering involves multiplying $\textbf{Y}_{lidar}^{2D}$ with the affinity matrix to produce ${\textbf{Y}_{lidar}^{2D}}'$:
\begin{equation}
  {\textbf{Y}_{lidar}^{2D}}'(i) = \frac{1}{z_i} \sum_{\forall j\in {\rm\Omega}} A^{cam}_{i,j}{\textbf{Y}_{lidar}^{2D}}(j),
  \label{eq_filtering}
\end{equation}
where $\rm\Omega$ is the set of pixels in $\textbf{Y}_{lidar}^{2D}$, 
and $z_i = \sum_{j}A_{i,j}$ is a normalization term. 
%
Note that, we adopt a linear-time algorithm~\cite{nips19_ltf} to implement the filtering, which reduces computational complexity from $\mathcal{O}((H\cdot W)^2)$ to $\mathcal{O}(H\cdot W)$.

By using the dense semantic relations from the camera image to constrain $\textbf{Y}_{lidar}^{2D}$, our guided filtering improves prediction accuracy in unlabeled regions.
However, the sparse property of the input LiDAR maps still challenges our task. 


\subsection{Dynamic Cross Pseudo Supervision}
\label{subsec:dycross}

Another challenge in achieving dense and accurate intermediate 2D predictions is the inherent sparsity of ${\textbf{Y}_{lidar}^{2D}}'$, which arises from the sparse nature of the input LiDAR maps. 
To address this, we introduce a dynamic cross pseudo supervision strategy to encourage ${\textbf{Y}_{lidar}^{2D}}'$ to emulate $\textbf{Y}_{cam}$, which inherits the dense distribution of camera images. 

To be specific, we formulate our strategy as a training loss: 
\begin{equation}
  \mathcal{L}_{dycross} = \mathcal{L}_{l2c}({\textbf{Y}_{lidar}^{2D}}'\parallel\textbf{Y}_{cam}) + \mathcal{L}_{c2l}(\textbf{Y}_{cam}\parallel{\textbf{Y}_{lidar}^{2D}}'),
  \label{eq_dycross}
\end{equation}
where 
$\mathcal{L}_{l2c}$ encourages ${\textbf{Y}_{lidar}^{2D}}'$ to emulate the dense distribution of $\textbf{Y}_{cam}$, and
$\mathcal{L}_{c2l}$ is an auxiliary term to ensure $\textbf{Y}_{cam}$ provides reliable supervision for ${\textbf{Y}_{lidar}^{2D}}'$.
We focus on the LiDAR-to-camera term to illustrate the mechanism of our strategy.
Specifically, we define $\mathcal{L}_{l2c}$ as: 
\begin{equation}
  \mathcal{L}_{l2c}({\textbf{Y}_{lidar}^{2D}}'\parallel\textbf{Y}_{cam}) = \frac{1}{HW}\sum_{h=1}^{H}\sum_{w=1}^{W} {W^{l2c}_{h,w}}D_{\rm KL} \big( {\textbf{Y}_{lidar}^{2D}}'(h,w)\parallel\textbf{Y}_{cam}(h,w) \big),
  \label{eq_l2c}
\end{equation}
where 
$W^{l2c}$ is a dynamic weight map, and $D_{\rm KL}(\cdot\parallel\cdot)$ represents the Kullback-Leibler (KL) divergence~\cite{pami22_xmuda,nipsw14_kl}. 
KL divergence helps align the distribution of ${\textbf{Y}_{lidar}^{2D}}'$ with that of $\textbf{Y}_{cam}$ by measuring their statistical distance. 
Since $\textbf{Y}_{cam}$ is a pseudo-supervision signal produced by our model and contains unreliable pixels, we set the dynamic weight map to emphasize reliable pixels: 
\begin{equation}
\begin{split}
\begin{aligned}
  W^{l2c}_{h,w} = \left \{
  \begin{array}{ll}
    C^{cam}_{h,w} - C^{lidar}_{h,w},~&{\rm if}~{C^{cam}_{h,w}> {\rm max}(C^{lidar}_{h,w},\tau)}, \\
    0,                         &{\rm otherwise},                         
  \end{array}
  \right.
  \label{eq_dynamic}
\end{aligned}
\end{split}
\end{equation}
where $C^{cam}\in\mathbb{R}^{H\times W}$ and $C^{lidar}\in\mathbb{R}^{H\times W}$ are the confidence maps~\cite{cvpr22_u2pl,iccv21_pmf} of $\textbf{Y}_{cam}$ and ${\textbf{Y}_{lidar}^{2D}}'$, respectively, and $\tau$ is a dynamic threshold that increases as training goes.  
By doing so, the LiDAR-to-camera term can effectively densify ${\textbf{Y}_{lidar}^{2D}}'$ by distilling the reliable, dense semantic knowledge from $\textbf{Y}_{cam}$. 
The camera-to-LiDAR term is formulated similarly to~\cref{eq_l2c}.

Combing our dynamic supervision loss with the focal and softmax losses, our full training objective becomes: 
\begin{equation}
  \mathcal{L} = \mathcal{L}_{foc}^{lidar} + \mathcal{L}_{lov}^{lidar} + \mathcal{L}_{foc}^{cam} + \mathcal{L}_{lov}^{cam} + \alpha\mathcal{L}_{dycross}.
  \label{eq_full}
\end{equation}

In~\cref{subsec:ablation}, we demonstrate that:  
(\uppercase\expandafter{\romannumeral1})~Our techniques enable intermediate 2D semantic predictions with dense distribution and high accuracy, despite the sparsity of input LiDAR maps and supervision label maps.
(\uppercase\expandafter{\romannumeral2})~Our improvements in 2D space effectively enhance the final 3D accuracy.


\section{Experiments}
\label{sec:experiments}

\subsection{Settings}
\label{subsec:settings}

\textbf{Datasets.}
We conduct experiments on nuScenes~\cite{cvpr20_nuscenes} and our nuScenes2D3D. 
nuScenes is a benchmark dataset for 3D LiDAR semantic segmentation. 
It provides 1,000 driving street scenes of 20 seconds, which are split into 28,130 training samples and 6,019 validation samples.  
Each sample includes 32-beam LiDAR point clouds and camera images from six views. 
16 semantic classes are adopted by merging similar and infrequent classes. 
Annotations are provided only for point clouds. 
%
Since nuScenes lacks 2D labels, which limits the evaluation of 2D semantic predictions, we introduce nuScenes2D3D, providing not only 3D labels on point clouds but also fine 2D labels on camera images. 
Our nuScenes2D3D includes 150 validation scenes and 150 training scenes sampled from nuScenes.
To lower redundancy and avoid class imbalance, we select the sample with the highest class diversity and instance count from each scene.
Each sample includes 3D semantic labels, 2D semantic labels for all six camera images, and bounding boxes and instance masks for foreground objects in the images.
In total, our nuScenes2D3D contains 300 labeled point clouds and 1,800 labeled camera images, with half as test data and half for training use. 
We adopt 16 classes for LiDAR segmentation.
More details are provided in the supplementary material.

\textbf{Network and Dynamic Cross Pseudo Supervision Loss.}
We use SalsaNext~\cite{isvc20_salsanext} as the LiDAR backbone and ResNet-34 or ResNet-50~\cite{cvpr16_resnet} as the camera backbone;
use a $1\times1$ convolution as the mapping layer to transform the encoder camera features; 
adopt an embedding layer of Conv$_{1\times1}$-BN-LeakyReLU to extract low-level camera feature. 
In $\mathcal{L}_{dycross}$, the threshold $\tau$ increases from 0.7 to 0.8 during training, with a loss weight $\alpha$ set to 0.5. 
Ablations on $\tau$ and $\alpha$ are provided in the supplementary material. 

\textbf{Implementation Details.} 
We follow the projection and remapping protocols in~\cite{iccv21_pmf}. 
During training, we minimize the objective in~\cref{eq_full}
using AdamW for the LiDAR backbone and SGD for the camera backbone. 
The learning rate starts at 0.001 and decays to 0 following a warmup cosine schedule. 
We apply basic augmentations, including random horizontal flipping, rotation, cropping, and color jitter. 
The model is trained for 50 epochs with a batch size of 64.
We adopt the above setting in all experiments\footnote{Since our task does not require dense 2D labels as supervision, for experiments on nuScenes2D3D, we train all methods using nuScenes training samples to fully utilize the available sparse labels.}. 
Evaluation is performed using mean Intersection over Union (mIoU).
We do not use any post-refinement such as CRF~\cite{nips11_crf} and kNN~\cite{iros19_rangenet++}.  


\subsection{Ablation Study}
\label{subsec:ablation}

\begin{table}[t]
  \centering
  {\small{
  \begin{tabular}{l|cccc}
    \hline
    & Model$_{1}$ (Baseline) & Model$_{2}$ & Model$_{3}$ & Model$_{4}$ (MM2D3D) \\   \hline
    Guided Filtering &\ding{55} &\textbf{\checkmark} &\ding{55}           &\textbf{\checkmark} \\
    Cross Supervision&\ding{55} &\ding{55}           &\textbf{\checkmark} &\textbf{\checkmark} \\ \hline
    2D mIoU (\%)     & 4.62     & 15.59              & 22.38              & \textcolor{blue}{45.61} \\
    3D mIoU (\%)     & 74.72    & 76.23              & 76.40              & \textcolor{blue}{77.53} \\
    \hline
  \end{tabular}
  }}
  \caption{
  Effectiveness study of cross-modal guided filtering and dynamic cross pseudo supervision.
  The best results are in \textcolor{blue}{blue}. 
  }
  \label{tab1_ablation_main}
\end{table}

\begin{figure}[t]
  \centering
  \begin{subfigure}{0.136\linewidth}
    \includegraphics[width=1\linewidth]{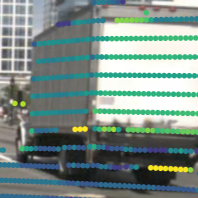}
    \caption{Inputs}  
  \end{subfigure}
  \begin{subfigure}{0.136\linewidth}
    \includegraphics[width=1\linewidth]{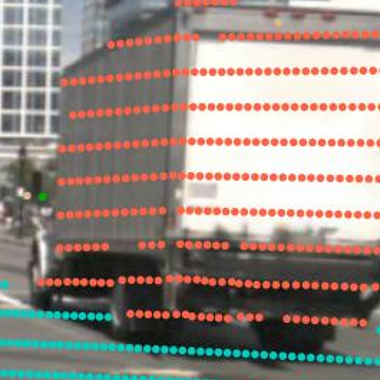}
    \caption{P3D Label}    
  \end{subfigure}
  \begin{subfigure}{0.136\linewidth}
    \includegraphics[width=1\linewidth]{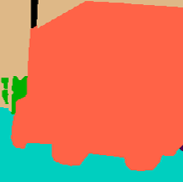}
    \caption{2D Label}   
  \end{subfigure} 
  \begin{subfigure}{0.136\linewidth}
    \includegraphics[width=1\linewidth]{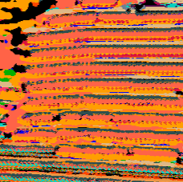}
    \caption{Model$_{1}$}
  \end{subfigure}
  \begin{subfigure}{0.136\linewidth}
    \includegraphics[width=1\linewidth]{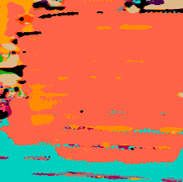}
    \caption{Model$_{2}$}
  \end{subfigure}
  \begin{subfigure}{0.136\linewidth}
    \includegraphics[width=1\linewidth]{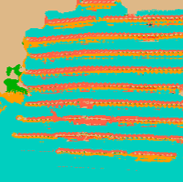}
    \caption{Model$_{3}$}
  \end{subfigure}
  \begin{subfigure}{0.136\linewidth}
    \includegraphics[width=1\linewidth]{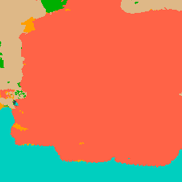}
    \caption{Model$_{4}$}
  \end{subfigure}
  \caption{ 
  Effect of our cross-modal guided filtering and dynamic cross pseudo supervision on intermediate 2D semantic predictions. 
  Only sparse label maps are used for training.
  }
  \label{fig4_ablation_main}
\end{figure}

We analyze our model by observing intermediate 2D predictions and final 3D accuracy on our nuScenes2D3D test set.
ResNet-34 is employed as the camera backbone.

\textbf{General Effectiveness.}
We first study the general effectiveness of the two introduced techniques in \cref{tab1_ablation_main}.
Compared to Model$_{1}$, a baseline that employs the architecture in~\cref{fig2_model} but removes our two techniques, 
Model$_{2}$ with our cross-modal guided filtering yields a $10.97\%$ improvement in 2D accuracy and a $1.51\%$ improvement in 3D accuracy.
Model$_{3}$ trained with our dynamic cross pseudo supervision strategy shows a $17.76\%$ gain in 2D accuracy and a $1.68\%$ gain in 3D accuracy.  
By jointly adopting our techniques,
Model$_{4}$, \textit{i.e.},~our MM2D3D, reaches 2D accuracy of $45.61\%$ and 3D accuracy of $77.53\%$, which are respectively $40.99\%$ and $2.81\%$ higher than the baseline. 
\Cref{fig4_ablation_main} visualizes the intermediate 2D semantic predictions from these four models.
In comparison with the baseline Model$_{1}$, Model$_{2}$ with cross-modal guided filtering increases accuracy in unlabeled regions by incorporating dense semantic relations from camera images but still shows some sparsity. 
Model$_{3}$, which uses dynamic cross pseudo supervision to encourage intermediate 2D predictions to emulate camera predictions, produces results with dense distribution but lower accuracy in unlabeled regions, such as the truck body.
By combining both techniques, Model$_{4}$ (our MM2D3D) achieves intermediate 2D semantic predictions that are dense and accurate, despite the sparsity of the input LiDAR maps and supervision labels.
Results in \cref{tab1_ablation_main} and \cref{fig4_ablation_main} demonstrate that shaping dense and accurate intermediate 2D predictions effectively enhances the final 3D accuracy. 


\textbf{Cross-Modal Guided Filtering.} 
Our guided filtering aims to leverage the dense semantics in camera images to guide intermediate 2D semantic predictions, which is not trivial due to spatial misalignment and information discrepancies between the camera images and LiDAR maps.
In~\cref{tab2_filtering}, we study different manners for learning our guided filtering.
\Cref{fig5_filtering} shows intermediate 2D semantic predictions from model variants. 
Model$_{5}$, which directly generates the tree affinity matrix from the camera image, produces 2D predictions that match the image but achieves low 3D accuracy.
This is due to misalignment between the camera image and the LiDAR map~(for example, see~\cref{fig5_filtering_input,fig5_filtering_p3dlabel}; points of the car may be projected onto the bodies of pedestrians), resulting in inaccurate 3D results when remapping the 2D predictions to the point cloud.  
Model$_{6}$, which generates tree affinities from the high-level feature of the camera image, achieves better 3D accuracy but lower 2D accuracy, as deep high-level features align better with the LiDAR map but lose important low-level structures and context.  
Model$_{4}$ generates affinities from the low-level feature of the camera image, achieving the highest 3D accuracy and superior 2D accuracy. 
This is because low-level features help mitigate misalignment while preserving detailed structural and contextual information.  
Model$_{7}$, which generates affinities from the low-level feature of the LiDAR map, exhibits lower 3D and 2D accuracy due to the lack of dense contextual information in the LiDAR map.  
%
Results in \cref{tab2_filtering} and \cref{fig5_filtering} show that leveraging low-level cross-modal semantic guidance addresses spatial misalignment and information discrepancy, effectively enhancing intermediate 2D predictions and final 3D accuracy.

\begin{table}[t]
  \centering
  {\small{
  \begin{tabular}{l|cccc}
    \hline
                    & Model$_{5}$             & Model$_{6}$ & Model$_{7}$ & Model$_{4}$ (MM2D3D) \\  
    \hline   
    Learning Manner & Cam-Image               & Cam-High    & LiDAR-Low   & Cam-Low \\ \hline
    2D mIoU (\%)    & \textcolor{blue}{56.21} & 40.02       & 10.49       & 45.61 \\
    3D mIoU (\%)    & 67.92                   & 76.29       & 76.23       & \textcolor{blue}{77.53} \\
  \hline
  \end{tabular}
  }}
  \caption{
  Study of different learning manners of our cross-modal guided filtering.
  The best results are in \textcolor{blue}{blue}. 
  }
  \label{tab2_filtering}
\end{table}

\begin{figure}[t]
  \centering
  \begin{subfigure}{0.136\linewidth}
    \includegraphics[width=1\linewidth]{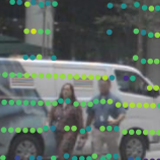}
    \caption{Inputs} 
    \label{fig5_filtering_input}
  \end{subfigure}
  \begin{subfigure}{0.136\linewidth}
    \includegraphics[width=1\linewidth]{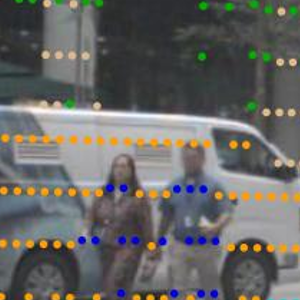}
    \caption{{P3D Label}}
    \label{fig5_filtering_p3dlabel}
  \end{subfigure}
  \begin{subfigure}{0.136\linewidth}
    \includegraphics[width=1\linewidth]{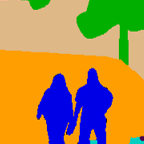}
    \caption{2D Label}
  \end{subfigure} 
  \begin{subfigure}{0.136\linewidth}
    \includegraphics[width=1\linewidth]{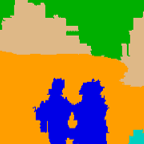}
    \caption{Model$_{5}$}
  \end{subfigure}
  \begin{subfigure}{0.136\linewidth}
    \includegraphics[width=1\linewidth]{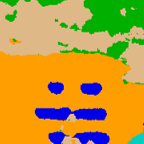}
    \caption{Model$_{6}$}
  \end{subfigure}
  \begin{subfigure}{0.136\linewidth}
    \includegraphics[width=1\linewidth]{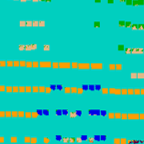}
    \caption{Model$_{7}$}
  \end{subfigure}
  \begin{subfigure}{0.136\linewidth}
    \includegraphics[width=1\linewidth]{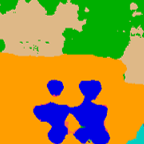}
    \caption{Model$_{4}$}
  \end{subfigure}
  \caption{
  Intermediate 2D semantic predictions from models employed different learning manners in cross-modal guided filtering. 
  Only sparse label maps are employed for training.
  }
  \label{fig5_filtering}
\end{figure}

\begin{table}[b]
  \centering
  {\small{
  \begin{tabular}{l|cccc}
    \hline
                & Model$_{8}$ & Model$_{9}$ & Model$_{10}$ & Model$_{4}$ (MM2D3D) \\   
    \hline
    Dynamic    & \ding{55}            & \ding{55}            & \textbf{\checkmark}   & \textbf{\checkmark}  \\
    Confidence & \ding{55}            & \textbf{\checkmark}  & \textbf{\checkmark}   & \textbf{\checkmark}  \\
    Cross      & \textbf{\checkmark}  & \textbf{\checkmark}  & \ding{55}             & \textbf{\checkmark}  \\   \hline
    2D mIoU (\%)     & 43.82    & 44.98  & 45.18   & \textcolor{blue}{45.61} \\  
    3D mIoU (\%)              & 75.55    & 77.08  & 77.38   & \textcolor{blue}{77.53} \\
  \hline       
  \end{tabular}
  }}
  \caption{
  Study of different components in our dynamic cross pseudo supervision strategy.
  The best results are in \textcolor{blue}{blue}. 
  }
  \label{tab3_dycross}
\end{table}


\textbf{Dynamic Cross Pseudo Supervision.} 
Our supervision strategy enhances intermediate 2D semantic predictions for LiDAR maps by dynamically transferring reliable semantic knowledge from camera predictions. 
\Cref{tab3_dycross} studies different components of our strategy. 
In Model$_{8}$, we retain the cross supervision manner in~\cref{eq_dycross} but remove the dynamic weight maps.
Due to lack of consideration for the reliability of intermediate 2D and camera semantic predictions, the 2D and 3D accuracy values decrease to $43.82\%$ and $75.55\%$, respectively. 
Based on Model$_{8}$, in Model$_{9}$, we set static weight maps by fixing the confidence threshold $\tau$ as 0.7.  
The accuracy values increase to $44.98\%$ and $77.08\%$, respectively. 
As prediction reliability increases during training, Model$_{4}$ (our MM2D3D) uses dynamic weight maps, achieving further improvements to $45.61\%$ for 2D prediction and $77.53\%$ for 3D accuracy. 
These improvements highlight the effectiveness of dynamically leveraging reliable prediction pixels.
In Model$_{10}$, we apply dynamic weight maps but remove the camera-to-LiDAR term in~\cref{eq_dycross}.
This results in decreased accuracy values, showing the necessity of including this auxiliary term to ensure that camera predictions provide reliable supervision for intermediate 2D predictions.


\begin{table}
  \centering
  {\small{
  \begin{tabular}{l|c|c|c|c}
    \hline
    Method                                & Inputs & Projection View & 2D mIoU (\%)  & 3D mIoU (\%) \\ \hline
    RangeNet++~\cite{iros19_rangenet++} & L & Range & - & 66.90 \\   
    SalsaNext~\cite{isvc20_salsanext}  & L & Range  & - & 71.32 \\ 
    FIDNet~\cite{iros21_fidnet}  & L & Range  & - & 72.08 \\   
    CENet~\cite{icme22_cenet} & L     & Range  & -     & 73.90 \\ 
    PMF-Res34~\cite{iccv21_pmf}                 & L+C   & Perspective & 29.52 & 76.37 \\
    PMF-Res50~\cite{iccv21_pmf} & L+C & Perspective & 32.01 & 77.95 \\
    RangeViT-IN21k~\cite{cvpr23_rangevit} & L     & Range  & - & 74.85 \\    
    RangeViT-CS~\cite{cvpr23_rangevit}    & L     & Range  & - & 75.48 \\ 
    EPMF-Res34~\cite{pami24_epmf} & L+C & Perspective  & - & 77.48 \\ 
    EPMF-Res50~\cite{pami24_epmf} & L+C & Perspective & - & 79.23 \\
    Baseline   & L+C   & Perspective & 4.62 & 74.72 \\  
    MM2D3D-Res34 (Ours) & L+C   & Perspective & 45.61 & 77.53 \\
    MM2D3D-Res50 (Ours) & L+C   & Perspective & \textcolor{blue}{49.22} & \textcolor{blue}{79.68} \\  
  \hline
  \end{tabular}
  }}
  \caption{
  Comparisons of projection-based LiDAR segmentation methods on our nuScenes2D3D test set.
  L denotes LiDAR point cloud, C denotes camera image. 
  Baseline is the Model$_{1}$ in~\cref{tab1_ablation_main}.
  All methods are trained with the same number of epochs and batch size for a fair comparison.  
  The best results are in \textcolor{blue}{blue}.} 
  \label{tab4_nuscenes2d3d}
\end{table}

\begin{figure}[!t]
  \centering
  \begin{subfigure}{0.16\linewidth}  
    \includegraphics[width=1\linewidth]{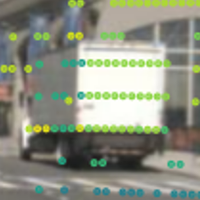}
    \caption{Inputs}
  \end{subfigure}
  \begin{subfigure}{0.16\linewidth}
    \includegraphics[width=1\linewidth]{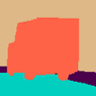}
    \caption{2D Label}   
  \end{subfigure}
  \begin{subfigure}{0.16\linewidth}  
    \includegraphics[width=1\linewidth]{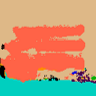}
    \caption{PMF~\cite{iccv21_pmf}}
  \end{subfigure} 
  \begin{subfigure}{0.155\linewidth}
    \includegraphics[width=1\linewidth]{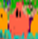}
    \caption{RViT-CS~\cite{cvpr23_rangevit}}
  \end{subfigure}
  \begin{subfigure}{0.16\linewidth}
    \includegraphics[width=1\linewidth]{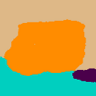}
    \caption{EPMF~\cite{pami24_epmf}}
  \end{subfigure}
  \begin{subfigure}{0.16\linewidth}
    \includegraphics[width=1\linewidth]{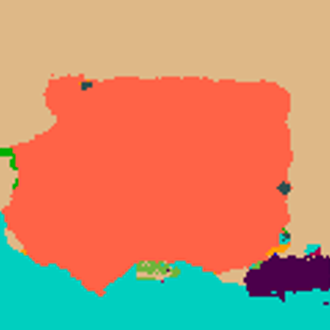}
    \caption{Ours} 
  \end{subfigure}   \\
  \begin{subfigure}{0.16\linewidth}    
    \includegraphics[width=1\linewidth]{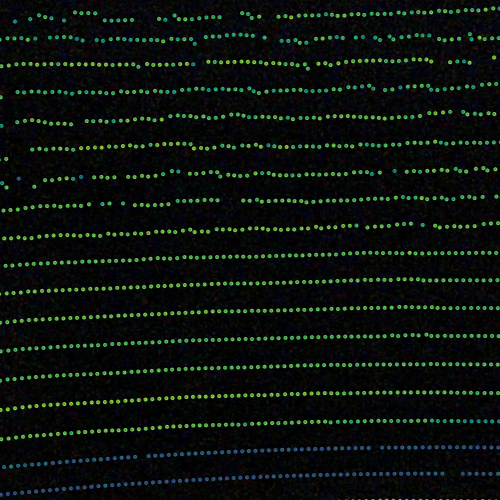}
    \caption{Inputs}
  \end{subfigure}
  \begin{subfigure}{0.16\linewidth}  
    \includegraphics[width=1\linewidth]{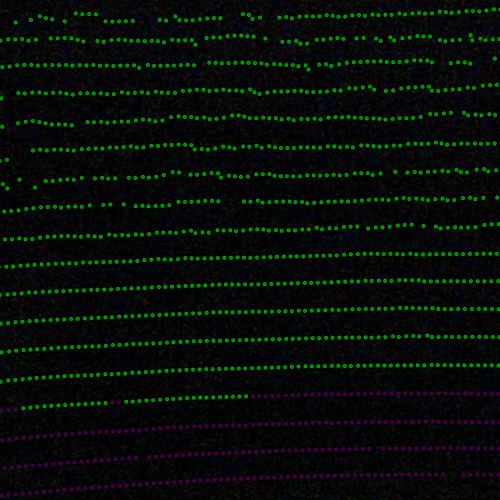}
    \caption{P3D Label} 
  \end{subfigure}
  \begin{subfigure}{0.16\linewidth}
    \includegraphics[width=1\linewidth]{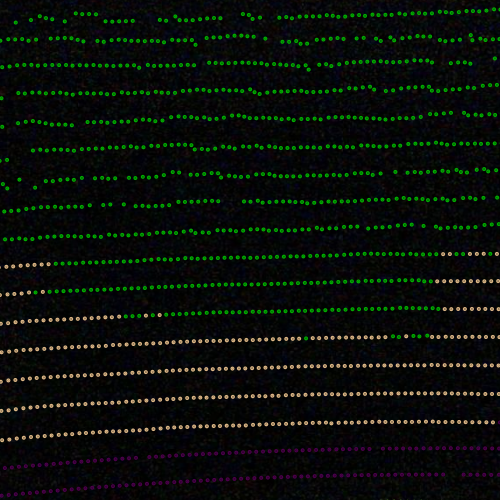}
    \caption{PMF~\cite{iccv21_pmf}}
  \end{subfigure} 
  \begin{subfigure}{0.16\linewidth}  
    \includegraphics[width=1\linewidth]{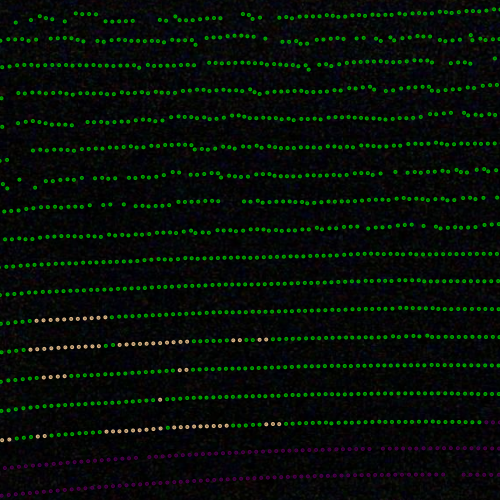}
    \caption{RViT-CS~\cite{cvpr23_rangevit}}
  \end{subfigure}
  \begin{subfigure}{0.16\linewidth}
    \includegraphics[width=1\linewidth]{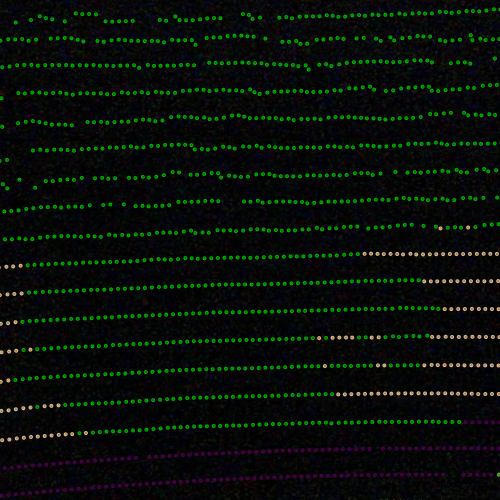}
    \caption{EPMF~\cite{pami24_epmf}}
  \end{subfigure}
  \begin{subfigure}{0.16\linewidth}  
    \includegraphics[width=1\linewidth]{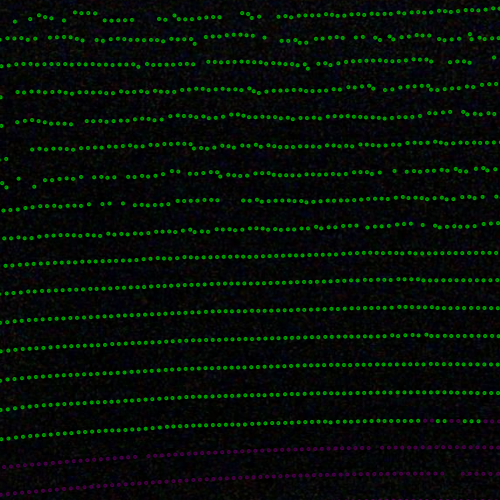}
    \caption{Ours}  
  \end{subfigure}
  \caption{
  Intermediate 2D semantic predictions (first two rows) 
  and projected 3D segmentation results (second two rows) on our nuScenes2D3D test set. 
  Note that the 2D prediction of RangeViT-CS is of range view, which has different coordinates from the perspective view; we enlarged it for better visual comparison.
  }
  \label{fig6_nuscenes2d3d}
\end{figure}

\subsection{Comparisons on nuScenes2D3D} 
\label{subsec:comparison_nuscenes2d3d}

We observe performance in both 2D and 3D spaces on our nuScenes2D3D test set.
\Cref{tab4_nuscenes2d3d} compares our model with state-of-the-art projection-based methods. 
Note that we do not compare with RangeFormer~\cite{iccv23_rangeformer} because its source code is not released; do not report the 2D accuracy of the range-view methods and EPMF because their intermediate 2D semantic predictions and the 2D labels are from different projection views or misaligned.
Our MM2D3D model achieves the best results in both 2D predictions and 3D accuracy.
\Cref{fig6_nuscenes2d3d} visualizes the intermediate 2D predictions and final 3D results for different methods. 
As shown in the first two rows, although RangeViT-CS benefits from transfer pretraining on image classification and segmentation tasks, and PMF leverages perception-aware multi-sensor fusion, both methods produce intermediate 2D semantic predictions that still exhibit sparsity and inaccuracy.
The very recent method EPMF also produces incorrect predictions.
In contrast, thanks to our cross-modal guided filtering and dynamic cross pseudo supervision, which effectively address intrinsic sparsity issues, our MM2D3D achieves intermediate 2D predictions with a dense distribution and more accurate class information.
Moreover, as shown in the last two rows, our model performs well even in challenging night scenes.
These quantitative and qualitative comparisons demonstrate the superiority of our MM2D3D model in both 2D and 3D spaces. 


\begin{table}[t]
  \centering
  {\small{
  \begin{tabular}{l|c|c|c|c}
    \hline
    \multirow{2}{*}{Method}  & \multirow{2}{*}{Inputs} & \multirow{2}{*}{Projection View} & Validation Set & Test Set \\  \cline{4-5}  
      & & & \multicolumn{2}{c}{3D mIoU (\%)} \\ \hline
    \multicolumn{1}{l|}{RangeNet++~\cite{iros19_rangenet++}} & L & Range & 65.5 & -  \\  
    \multicolumn{1}{l|}{SalsaNext~\cite{isvc20_salsanext}} & L & Range & 72.2 & -  \\  
    \multicolumn{1}{l|}{FIDNet~\cite{iros21_fidnet}} & L & Range & 71.4 & 72.8 \\  
    \multicolumn{1}{l|}{CENet~\cite{icme22_cenet}} & L & Range & 73.3 & 74.7 \\
    \multicolumn{1}{l|}{PMF-Res34~\cite{iccv21_pmf}} & L+C & Perspective & 76.9 & 75.5 \\ 
    \multicolumn{1}{l|}{PMF-Res50~\cite{iccv21_pmf}} & L+C & Perspective & 79.0 & 77.0  \\ 
    \multicolumn{1}{l|}{RangeViT-IN21k~\cite{cvpr23_rangevit}} & L & Range & 74.8 & 73.6$^{\dag}$ \\   
    \multicolumn{1}{l|}{RangeViT-CS~\cite{cvpr23_rangevit}} & L & Range & 75.2 & 74.7$^{\dag}$ \\ 
    \multicolumn{1}{l|}{RangeFormer~\cite{iccv23_rangeformer}} & L & Range & 78.1 & 80.1 \\
    \multicolumn{1}{l|}{EPMF-Res34$^{\dag}$~\cite{pami24_epmf}} & L+C & Perspective & 78.9 & 77.8 \\ 
    \multicolumn{1}{l|}{EPMF-Res50$^{\dag}$~\cite{pami24_epmf}} & L+C & Perspective & \textcolor{blue}{80.2}  & 79.0  \\
    \multicolumn{1}{l|}{MM2D3D-Res34 (Ours)}  & L+C & Perspective  & 77.4 & 78.4  \\ 
    \multicolumn{1}{l|}{MM2D3D-Res50 (Ours)}  & L+C & Perspective  & 80.0 & \textcolor{blue}{80.3}  \\  
  \hline
  \end{tabular}
  }}
  \caption{
  Comparisons of projection-based methods on the nuScenes validation and test sets.~L denotes LiDAR point cloud, C denotes camera image.
  $^{\dag}$Our implementation; Trained with the same number of epochs and batch size as ours for a fair comparison.
  The best results are in \textcolor{blue}{blue}.  
  }
  \label{tab5_nuscenes}
\end{table}

\begin{figure}[t]
  \centering
  \begin{subfigure}{0.25\linewidth}  
    \includegraphics[width=1\linewidth]{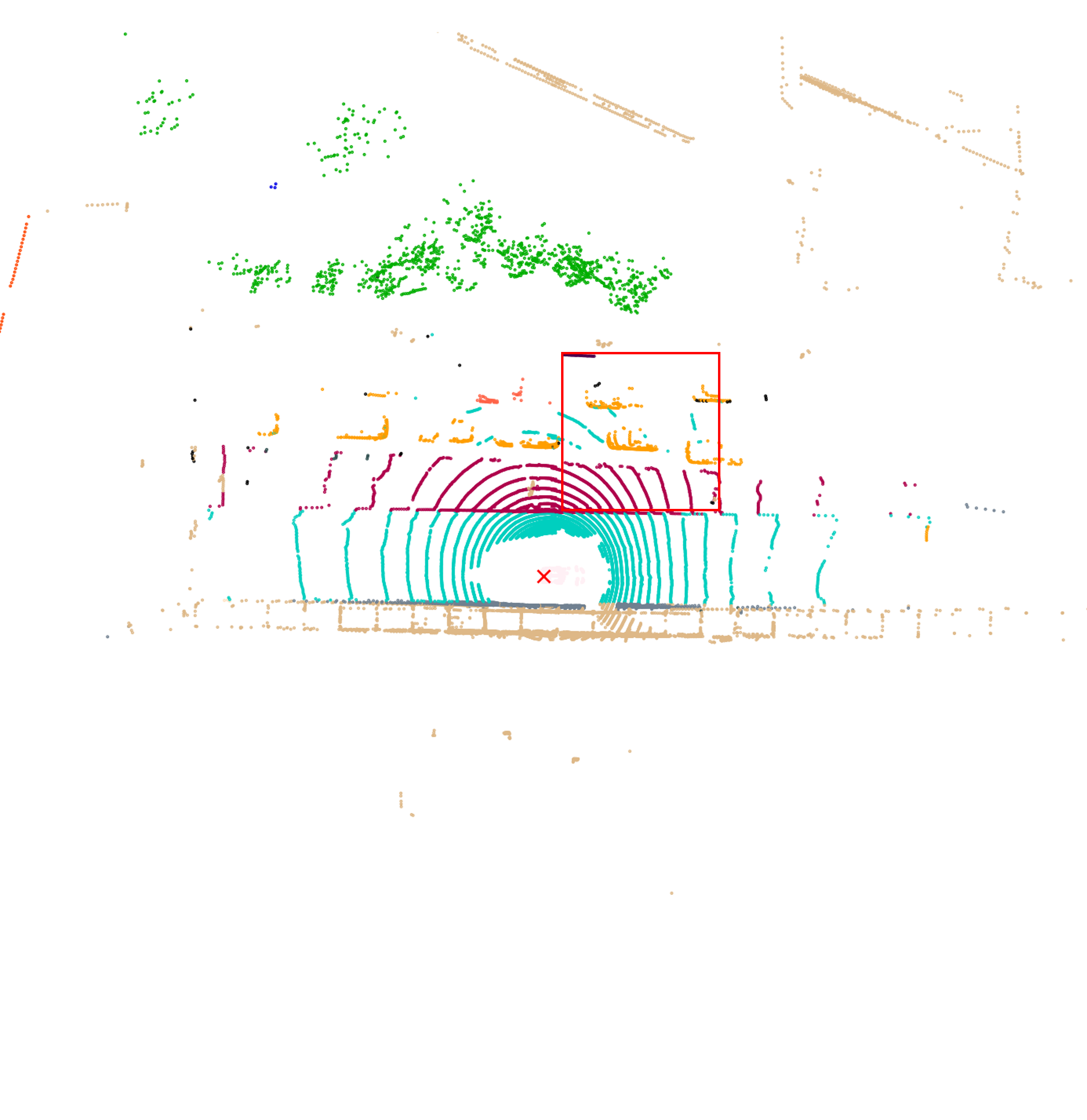}
    \centering{\scriptsize Ground Truth}
  \end{subfigure}
  \begin{subfigure}{0.11\linewidth} 
    \includegraphics[width=1\linewidth]{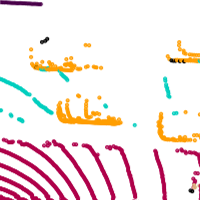}
    {\vspace{0.14cm} \centering{\scriptsize GT}}      
    \includegraphics[width=1\linewidth]{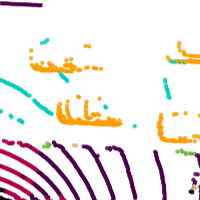}
    \centering{\scriptsize EPMF~\cite{pami24_epmf}} 
  \end{subfigure} 
  \begin{subfigure}{0.11\linewidth}
    \includegraphics[width=1\linewidth]{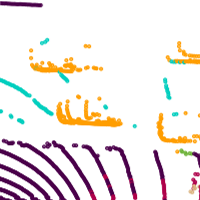}
    {\vspace{0.08cm} \centering{\scriptsize RViT-CS~\cite{cvpr23_rangevit}}}  
    \includegraphics[width=1\linewidth]{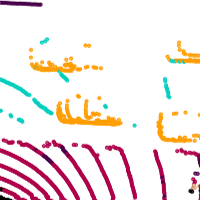}
    \centering{\scriptsize Ours}  
  \end{subfigure}  
  {~}
  \begin{subfigure}{0.25\linewidth}   
    \includegraphics[width=1\linewidth]{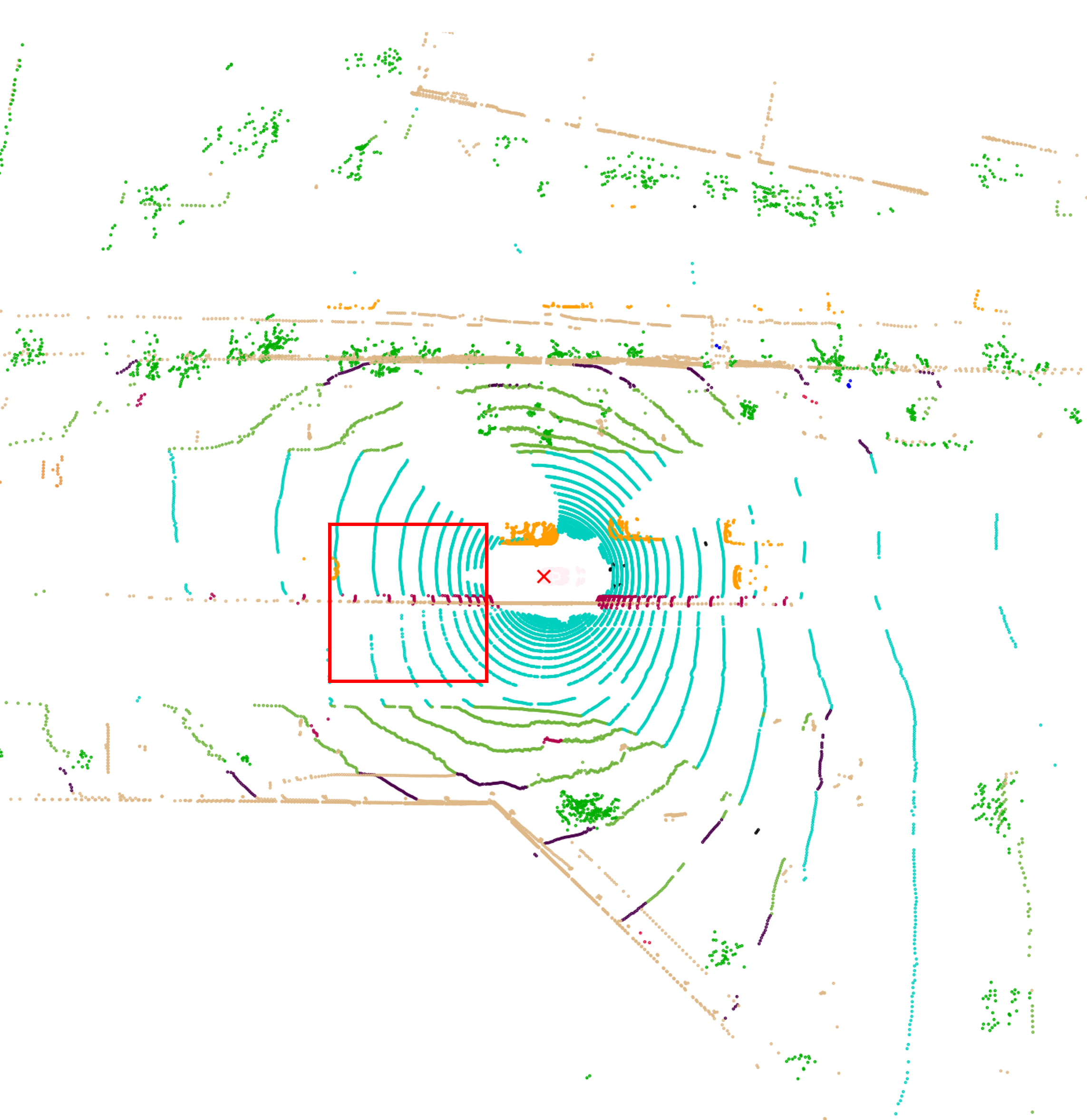}
    \centering{\scriptsize Ground Truth}
  \end{subfigure}
  \begin{subfigure}{0.11\linewidth} 
    \includegraphics[width=1\linewidth]{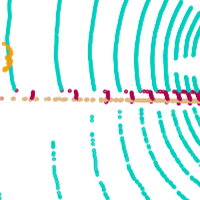}
    {\vspace{0.14cm} \centering{\scriptsize GT}}      
    \includegraphics[width=1\linewidth]{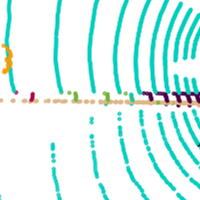}
    \centering{\scriptsize EPMF~\cite{pami24_epmf}}
  \end{subfigure} 
  \begin{subfigure}{0.11\linewidth}
    \includegraphics[width=1\linewidth]{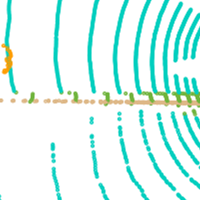}
    {\vspace{0.08cm} \centering{\scriptsize RViT-CS~\cite{cvpr23_rangevit}}}  
    \includegraphics[width=1\linewidth]{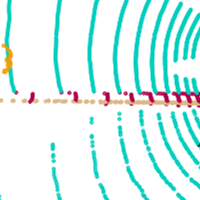}
    \centering{\scriptsize Ours}
  \end{subfigure} \\
  \caption{
  3D segmentation results on the nuScenes validation set.
  }
  \label{fig7_nuscenes}
\end{figure}

\subsection{Comparisons on nuScenes} 
\label{subsec:comparison_nuscenes}

We also validate our model on the nuScenes validation and test sets. 
\Cref{tab5_nuscenes} shows the quantitative results, with only 3D accuracy reported since nuScenes does not provide 2D semantic labels.
Our model outperforms previous projection-based methods, such as RangeFormer, RangeViT-CS, and PMF, on both the nuScenes validation and test sets, narrowing the gap with the very recent EPMF. 
presents visualizations of 3D segmentation results from different methods.
While both EPMF and RangeViT-CS produce incorrect segmentations, our MM2D3D provides more accurate class predictions and achieves results closer to the ground truth.
We provide more qualitative results in the supplementary material. 


\begin{figure}[t]
  \centering
  \begin{subfigure}{0.16\linewidth}   
    \includegraphics[width=1\linewidth]{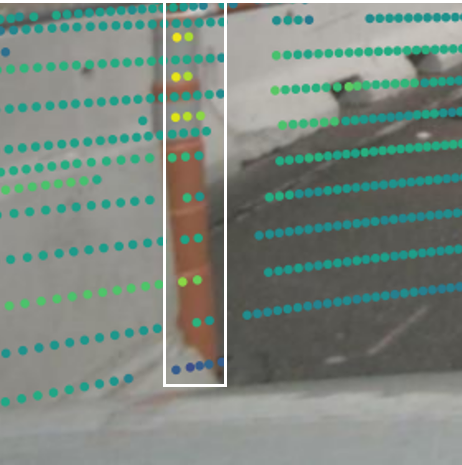}
    \caption{Inputs} %
    \label{cone_input}
  \end{subfigure}
  \begin{subfigure}{0.16\linewidth}
    \includegraphics[width=1\linewidth]{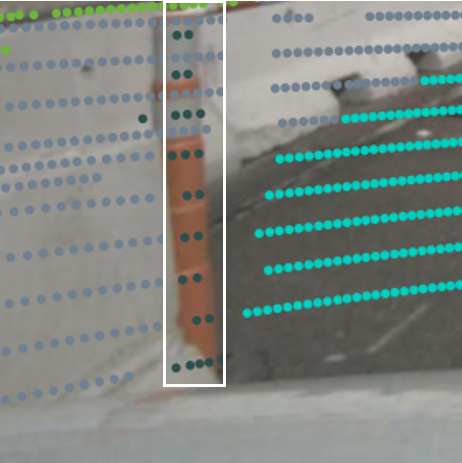}
    \caption{P3D Label}  %
    \label{cone_label_proj}
  \end{subfigure}
  \begin{subfigure}{0.16\linewidth}
    \includegraphics[width=1\linewidth]{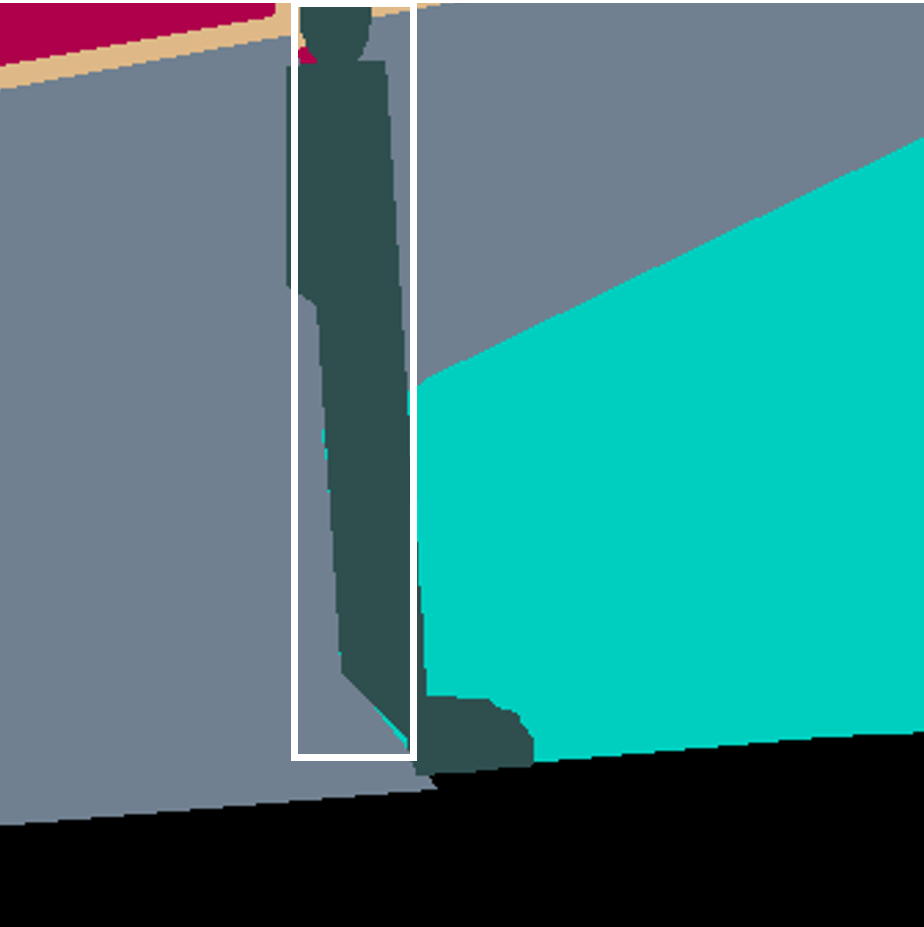}
    \caption{2D Label}  %
  \end{subfigure} 
  \begin{subfigure}{0.16\linewidth}  
    \includegraphics[width=1\linewidth]{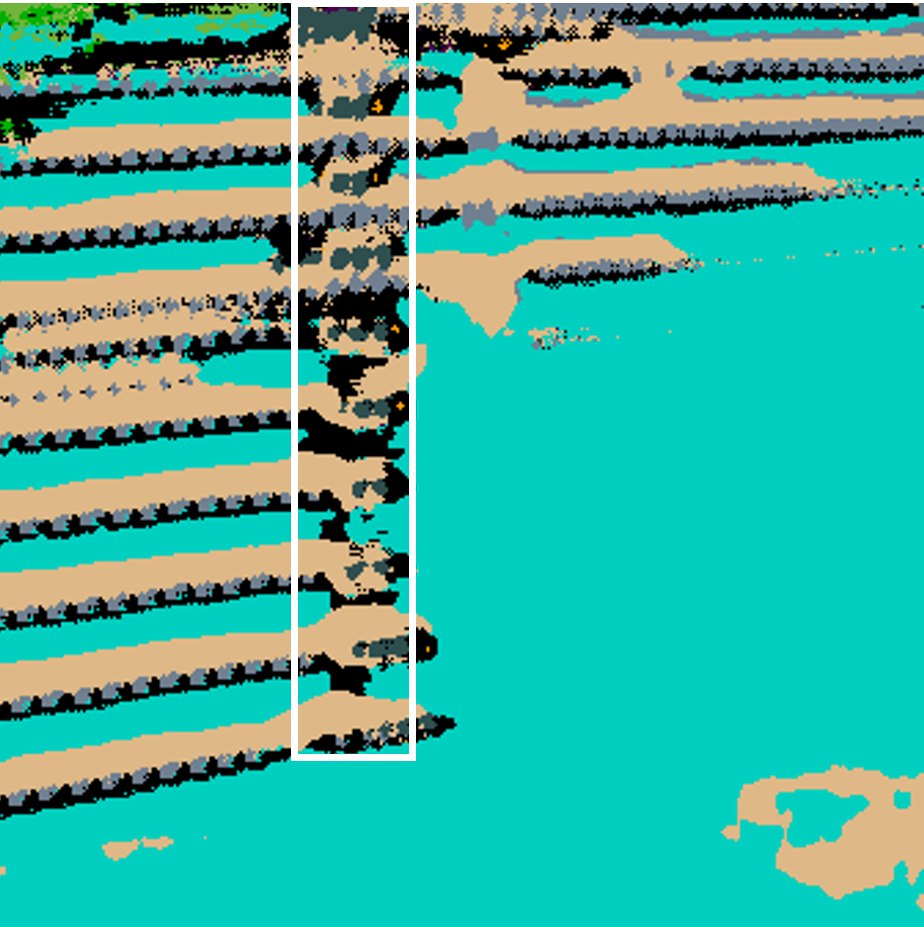}
    \caption{PMF~\cite{iccv21_pmf}}  %
  \end{subfigure}
  \begin{subfigure}{0.16\linewidth}  
    \includegraphics[width=1\linewidth]{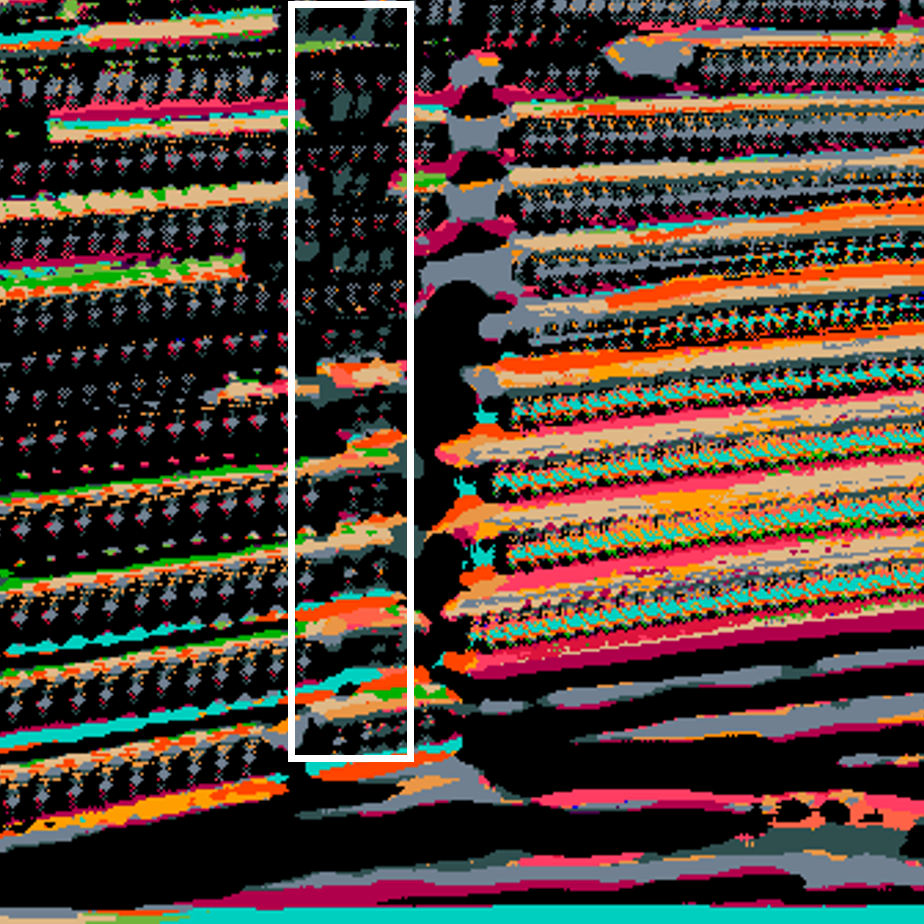}
    \caption{Baseline}  %
  \end{subfigure}
  \begin{subfigure}{0.16\linewidth}
    \includegraphics[width=1\linewidth]{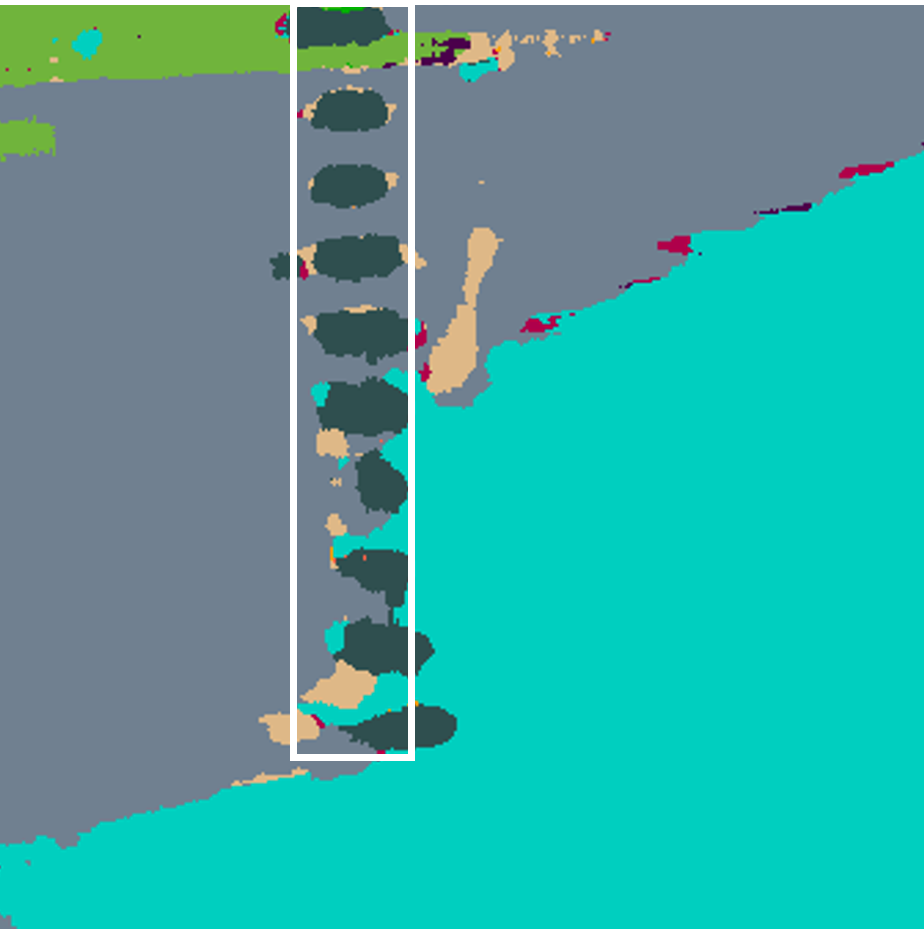}
    \caption{Ours} 
    \label{cone_ours}
  \end{subfigure} 
  \begin{subfigure}{0.16\linewidth}   
    \includegraphics[width=1\linewidth]{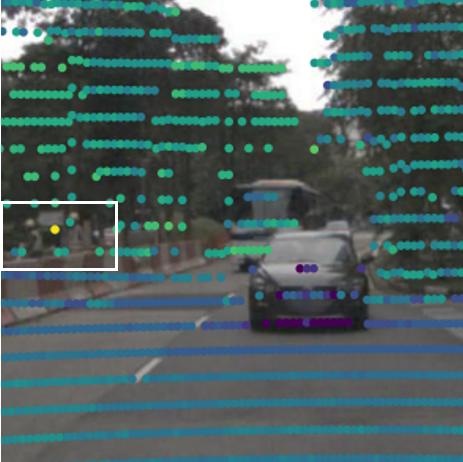}
    \caption{Inputs}
    \label{pedestrian_input}
  \end{subfigure}
  \begin{subfigure}{0.16\linewidth}
        \includegraphics[width=1\linewidth]{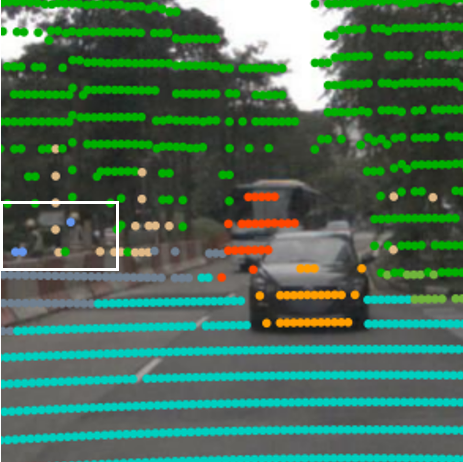}
    \caption{P3D Label}  
    \label{pedestrian_label_proj}
  \end{subfigure}
  \begin{subfigure}{0.16\linewidth}
    \includegraphics[width=1\linewidth]{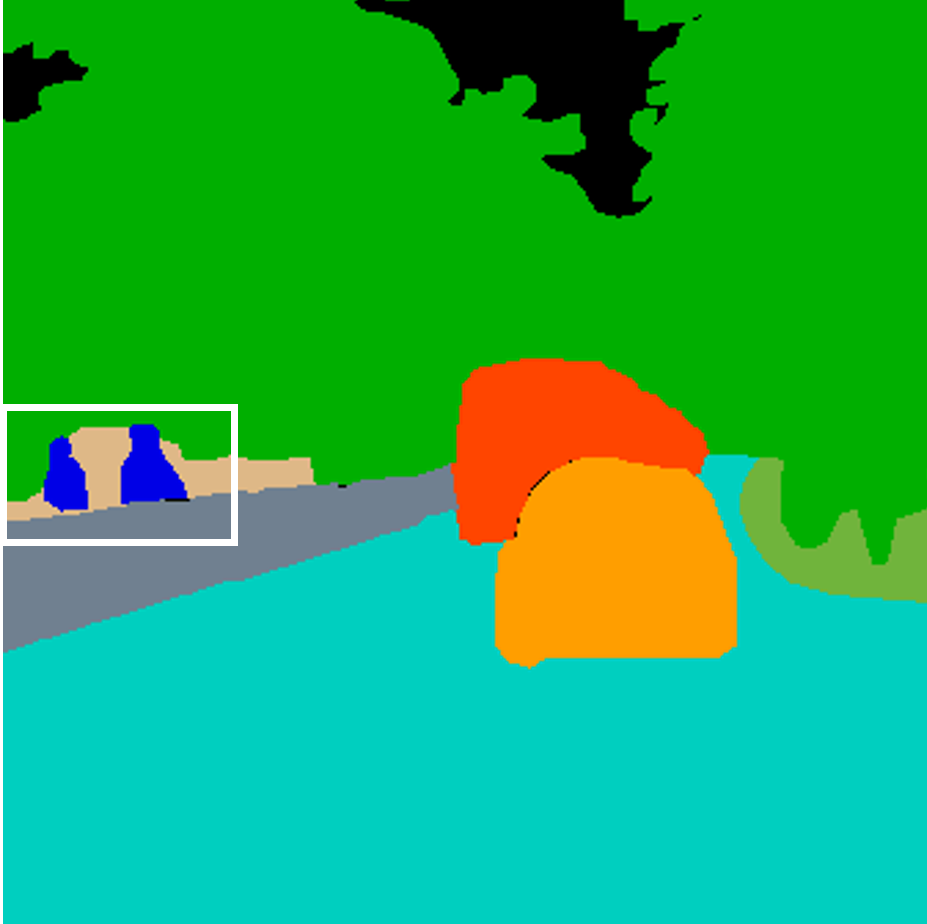}
    \caption{2D Label}  
  \end{subfigure}  
  \begin{subfigure}{0.16\linewidth}  
    \includegraphics[width=1\linewidth]{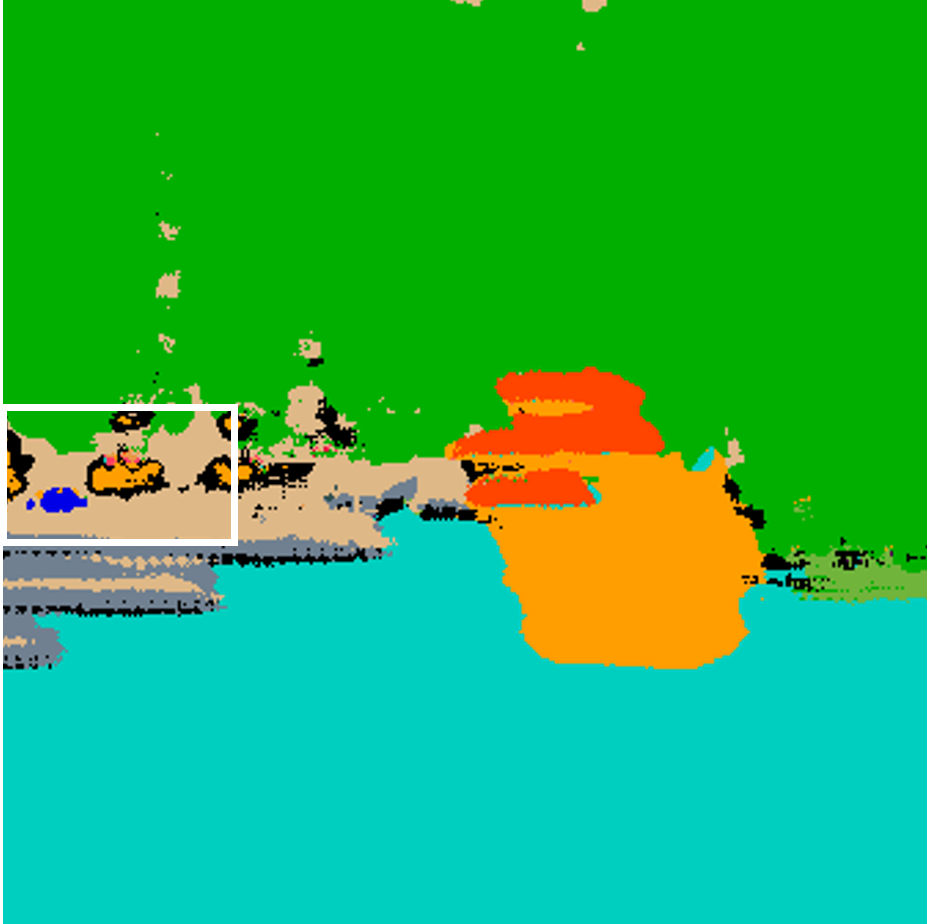}
    \caption{PMF~\cite{iccv21_pmf}} 
  \end{subfigure}
  \begin{subfigure}{0.16\linewidth}  
    \includegraphics[width=1\linewidth]{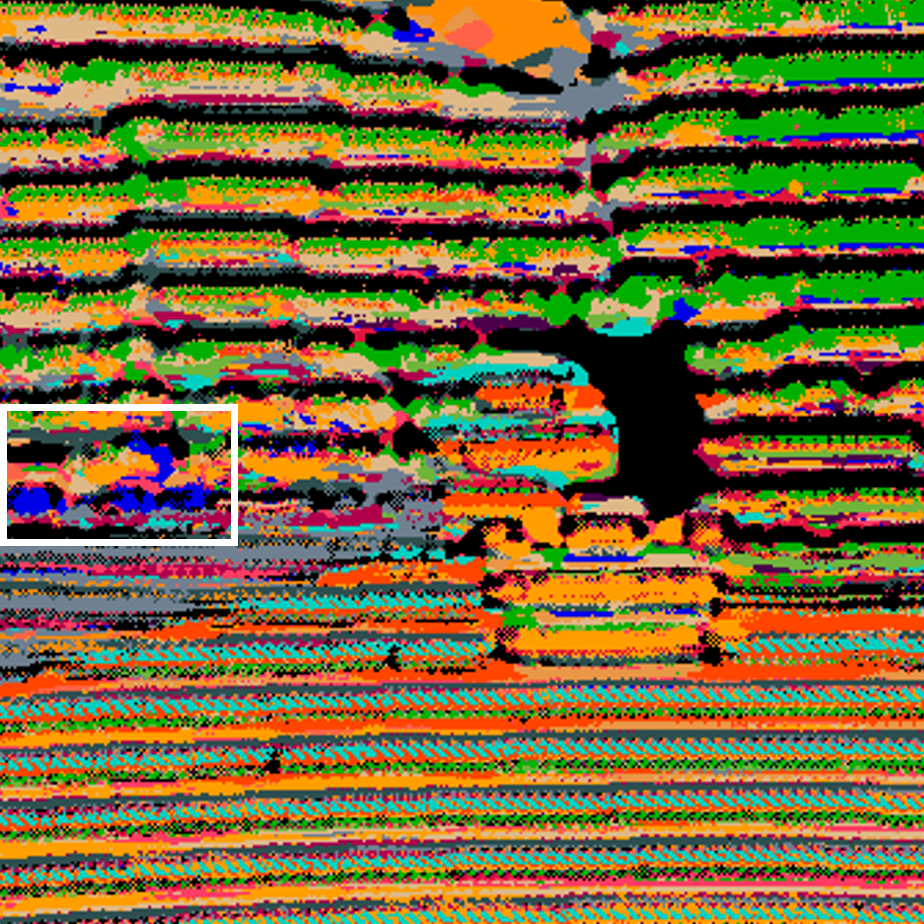}
    \caption{Baseline}
  \end{subfigure}
  \begin{subfigure}{0.16\linewidth}
    \includegraphics[width=1\linewidth]{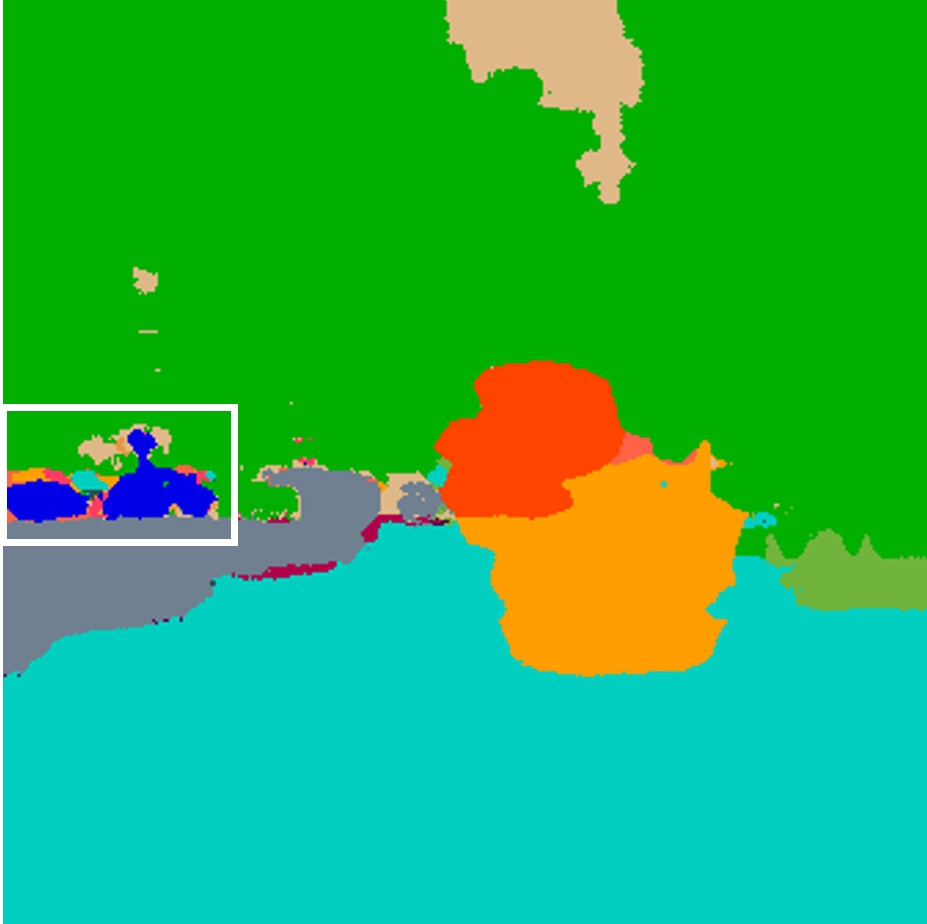}
    \caption{Ours}
    \label{pedestrian_ours}
  \end{subfigure} 
  \caption{
  Incomplete intermediate 2D semantic predictions on regions with few points.
  Only sparse labels are used for training. 
  }
  \label{fig8_failure}
\end{figure}

\section{Limitation}
\label{sec:limitation}

The adoption of our cross-modal guided filtering and dynamic cross pseudo supervision strategy effectively enhances projection-based 3D LiDAR segmentation by achieving dense and accurate intermediate 2D semantic predictions.
However, we observe failure cases in regions with few points. 
For instance, our model produces incomplete predictions on thin traffic cones (\cref{cone_ours}) and distant pedestrians (\cref{pedestrian_ours}). 
This is due to input LiDAR points and supervision label points for these objects are too few (see~\cref{cone_input,cone_label_proj,pedestrian_input,pedestrian_label_proj}).
But even so, our model still delivers denser and more accurate intermediate 2D predictions compared to others and achieves promising 3D segmentation performance. 
%
Another limitation is that our techniques require camera images as auxiliary data.
An interesting future direction would be enhancing LiDAR segmentation by exploring unsupervised depth completion~\cite{Wong_2021_ICCV}.


\section{Conclusion}
We enhance 3D LiDAR segmentation in the projection-based setting by shaping dense and accurate 2D semantic predictions.
By leveraging camera images as auxiliary data, we develop the MM2D3D model, which leverages cross-modal semantic guidance and encourages dynamic cross-mimicking to address the intrinsic sparsity issues in our task.
Experiments show that our techniques lead to intermediate 2D semantic predictions with dense distribution and higher accuracy, which effectively enhances the final 3D segmentation results. 
Comparisons with previous methods demonstrate our superiority in both 2D and 3D spaces.


\section{Acknowledgments}
This work was supported by the RIKEN Junior Research Associate (JRA) Program and JST FOREST Program Grant Number JPMJFR206S.








\end{document}